%% file: LNUCB-TA_JMLR.tex
\documentclass[twoside,11pt]{article}

\usepackage{jmlr2e}
\input{math_commands.tex}

\usepackage{graphicx} 
\usepackage{longtable}
\usepackage{hyperref}
\usepackage{url}
\usepackage{booktabs}
\usepackage{nicefrac}
\usepackage{xcolor}
\usepackage{algorithm}
\usepackage{algpseudocode}
\usepackage{float}
\usepackage{soul}
\usepackage{tabularx}
\usepackage{amsmath}
\usepackage{amsfonts}
\usepackage{makecell}
\usepackage{array}
\usepackage{parskip}
\usepackage{pifont}

\newcommand{\eg}{e.g.}
\newcommand{\ie}{i.e.}

\usepackage{lastpage}

\ShortHeadings{LNUCB-TA: Linear-nonlinear Hybrid Bandit Learning with Temporal Attention}{Khosravi, Shafie, Raihan, Das, Ahmed}
\firstpageno{1}

\begin{document}
	
	\title{LNUCB-TA: Linear-nonlinear Hybrid Bandit Learning with Temporal Attention}
	
	\author{\name Hamed Khosravi \email hamed.khosravi@mail.wvu.edu \\
		\addr Department of Industrial \& Management Systems Engineering\\
		West Virginia University\\
		Morgantown, WV 26505, USA
		\AND
		\name Mohammad Reza Shafie \email mr.shafie7731@gmail.com \\
		\addr Department of Electrical Engineering \\ 
		Iran University of Science and Technology\\
		Tehran, Tehran, Iran 
		\AND
		\name Ahmed Shoyeb Raihan \email ar00065@mix.wvu.edu \\
		\addr Department of Industrial \& Management Systems Engineering\\
		West Virginia University\\
		Morgantown, WV 26505, USA
		\AND
		\name Srinjoy Das\footnotemark[1] \email sd00052@mix.wvu.edu \\
		\addr School of Mathematical \& Data Sciences\\
		West Virginia University\\
		Morgantown, WV 26505, USA
		\AND
	\name Imtiaz Ahmed\footnotemark[1] \email imtiaz.ahmed@mail.wvu.edu \\
	\addr Department of Industrial \& Management Systems Engineering\\
	West Virginia University\\
	Morgantown, WV 26505, USA
}

\footnotetext[1]{\textsuperscript{*}Corresponding author.}
	\editor{....}
	
	\maketitle
	
	\begin{abstract}
		Existing contextual multi-armed bandit (MAB) algorithms fail to effectively capture both long-term trends and local patterns across all arms, leading to suboptimal performance in environments with rapidly changing reward structures. They also rely on static exploration rates, which do not dynamically adjust to changing conditions. To overcome these limitations, we propose LNUCB-TA, a hybrid bandit model integrating a novel nonlinear component (adaptive $k$-Nearest Neighbors ($k$-NN)) for reducing time complexity, alongside a global-and-local attention-based exploration mechanism. Our approach uniquely combines linear and nonlinear estimation techniques, with the nonlinear module dynamically adjusting $k$ based on reward variance to enhance spatiotemporal pattern recognition. This reduces the likelihood of selecting suboptimal arms while improving reward estimation accuracy and computational efficiency. The attention-based mechanism ranks arms by past performance and selection frequency, dynamically adjusting exploration and exploitation in real time without requiring manual tuning of exploration rates. By integrating global attention (assessing all arms collectively) and local attention (focusing on individual arms), LNUCB-TA efficiently adapts to temporal and spatial complexities. Empirical results show LNUCB-TA significantly outperforms state-of-the-art linear, nonlinear, and hybrid bandits in cumulative and mean reward, convergence, and robustness across different exploration rates. Theoretical analysis further confirms its reliability with a sub-linear regret bound.
	\end{abstract}
	
	
	\section{Introduction}

	The multi-armed bandit (MAB) problem brings to light a fundamental challenge in decision-making dynamics, emphasizing the need to strike balance between exploration and exploitation \citep{russac2019weighted, audibert2009exploration, hillel2013distributed}. In Reinforcement Learning (RL), this challenge manifests as a continuous decision-making process \citep{zhu2022flexible}. Specifically, the RL agents must navigate the trade-off between uncovering new opportunities to better exploit their environment versus leveraging proven strategies to realize immediate benefits \citep{reeve2018k, bouneffouf2020survey, sani2012risk}. Balancing this trade-off is critical for developing adaptive strategies to improve outcomes across various domains such as online advertising \citep{schwartz2017customer}, recommendation systems \citep{li2010contextual, ding2021hybrid}, and clinical trials \citep{villar2015multi, aziz2021multi}. This dilemma becomes pronounced in environments marked by uncertainty, \eg, digital marketing \citep{shi2023deep}. In these scenarios, algorithms aim to maximize user engagement by deciding advertisements displays to different segments, \ie, weighing the benefits of exploring diverse advertisements against exploiting those with proven success.

	\paragraph{Foundational approaches.}
	Given the extensive literature on MAB, our study specifically concentrates on Upper Confidence Bound (UCB) variants and linear estimation methods. Foundational methods such as the UCB algorithm optimize decision-making by constructing confidence bounds around estimated rewards and selecting the action with the highest upper bound \citep{auer2002finite}. This technique is further refined in the Kullback-Leibler Upper Confidence Bound (KL-UCB) algorithm, which enhances the accuracy of these intervals using the Kullback-Leibler divergence \citep{garivier2011kl}. Despite their efficacy, both UCB and KL-UCB often overlook the crucial role of contextual information, where each action can be tailored to the specific observable environmental factors, or `contexts' to maximize the obtained rewards \citep{bubeck2012regret}.
	
	\noindent Extending these concepts to address contextual dynamics, the Linear Upper Confidence Bound (LinUCB) algorithm assumes a linear relationship between contextual features and expected rewards \citep{chu2011contextual, dimakopoulou2019balanced}. LinUCB constructs confidence bounds around these estimated rewards and selects actions based on the upper bounds of these estimates \citep{li2010contextual}. Linear Thompson Sampling (LinThompson) also operates under the assumption that expected rewards are linearly related to contextual features, utilizing Thompson Sampling (TS) to balance exploration and exploitation\citep{agrawal2013thompson}. Despite its strategic approach, LinThompson can fall short by often estimating influence probabilities directly, which can lead to locally optimal solutions due to insufficient exploration. To address this, the LinThompsonUCB algorithm combines linear estimation with TS’s probabilistic approach and UCB confidence intervals to enhance exploration and performance. \citep{zhang2019automatic}. However, while effective, the reliance of LinUCB, LinThompson, and LinThompsonUCB on linear assumptions can limit their performance in more complex environments. To address this limitation, the $k$-Nearest Neighbour UCB ($k$-NN UCB) and $k$-Nearest Neighbour KL-UCB ($k$-NN KL-UCB) methods exploit the locality of feature space to enhance action selection \citep{reeve2018k}. 
	
	Recent advances in neural networks have led to new approaches that transcend linear assumptions. Early work on NeuralBandit introduced neural networks to estimate reward probabilities without linear separability constraints, employing stochastic gradient descent for online non-stationary adaptation \citep{allesiardo2014neural}. Subsequent innovations like NeuralUCB \citep{zhou2020neural} improved exploration efficiency by replacing $\epsilon$-greedy strategies with UCB-based mechanisms while maintaining a single network architecture. The NeuralTS framework \citep{zhang2021neural} advanced Thompson sampling by incorporating full network uncertainty through gradient-derived feature mappings, contrasting with hybrid approaches like NeuralLinear that apply Bayesian linear regression only to final-layer representations \citep{riquelme2018neural}. The Neural Contextual Bandits without Regret framework \citep{kassraie2022neural} further strengthens theoretical guarantees, establishing sublinear regret bounds via neural tangent kernel analysis.

    \paragraph{Existing gaps and intuition.} Despite advancements in MAB algorithms, existing algorithms predominantly fail to incorporate adaptive strategies for reward estimation as a function of the context. Linear models, constrained by static parameter updates, often fail in scenarios with inherently nonlinear relationships between contextual features and rewards, leading to outdated estimations and slower convergence \citep{russac2019weighted,dimakopoulou2019balanced,zhang2019automatic}. While nonlinear approaches like $k$-NN-based models  \citep{reeve2018k} offer flexibility, they often struggle with computational efficiency and adaptability in dynamic environments. Moreover, these models usually overlook crucial long-term trends, which can lead to overfitting in sparse scenarios, degraded generalization, and increased variance in reward estimations \citep{eleftheriadis2024empirical}. These limitations restrict existing algorithms' ability to capture both long-term trends and immediate local patterns effectively, leading to inconsistent performance across various scenarios.
    
    Practical concerns about the computational cost in exploration are even more significant for neural bandits like NeuralUCB and NeuralTS. These models require the construction of high-probability confidence sets based on the dimensionality of network parameters and context vector representations, often involving matrices with hundreds of thousands of parameters. As a result, approximations (e.g., only using diagonal covariance matrices) are employed to mitigate this computational burden \citep{kassraie2022neural,zhou2020neural}, but these approximations lack theoretical guarantees, creating gaps between theoretical and empirical performance \citep{jia2022learning}. The Neural bandit with perturbed reward (NPR) model \citep{jia2022learning} attempts to address computational efficiency in neural contextual bandits but highlights that online model updates in neural bandit models, relying on stochastic gradient descent over entire training sets at each round, remain a significant computational bottleneck \citep{goktas2024efficient,lee2024soft}.
    
    In addition, conventional methods rely on static exploration rates, leading to inefficient convergence and suboptimal decision-making \citep{bubeck2012regret}. Specifically, high exploration rates cause algorithms to frequently test suboptimal options, slowing progress and increasing regret \citep{audibert2009exploration}. Conversely, low exploration rates lead to premature conclusions on less optimal solutions, foregoing potentially better options \citep{odeyomi2020learning}. To address this, studies have proposed fine-tuning, experimentation, and dynamic exploration rates \citep{carlsson2021thompson, russac2019weighted,alon2015online}. However, these approaches often fall short in fully capturing the intricate, evolving patterns of rewards in non-stationary environments, such as recommendation systems or clinical trials \citep{villar2015multi, liu2024learning, de2023llm}. Existing solutions typically rely on pre-defined heuristics or manual tuning, which can be suboptimal when rewards shift unexpectedly, complicating the search for an optimal setting \citep{bouneffouf2020survey, russac2019weighted}. A key challenge of the existing approaches is to effectively adapt exploration rates as reward distributions change over time. As a result, context-awareness becomes critical to successfully manage exploration \citep{liu2024learning}.

	\paragraph{Contribution.}
	In this work, we have developed LNUCB-TA, which introduces a novel nonlinear strategy through an adaptive {$k$}-NN that dynamically adapts based on reward characteristics and shifts, effectively solving the time complexity issues commonly associated with nonlinear models. It also presents an attention-based exploration factor to move beyond the constraints of existing exploration rates. This model culminates in a unique synthesis of linear and nonlinear hybrid contextual MAB algorithms, comprehensively addressing the need for adaptive strategies in reward estimation to simultaneously capture long-term trends as well as local patterns across all arms. As shown in Table~\ref{comparison-table-CONTRIBUTION}, LNUCB-TA incorporates a linear component for a global approximation of the reward function and a unique nonlinear component for capturing local patterns. The proposed nonlinear component employs a data-driven (variance-guided), non-parametric criterion for \(k\) selection based on reward histories to reduce time complexity. Complementing this, the attention-based mechanism, inspired by the global-and-local attention (GALA) concept \citep{linsley2018learning}, dynamically adjusts the exploration strategy by utilizing past interactions and rewards. This temporal attention approach adaptively prioritizes arms based on their historical rewards and selection frequency, eliminating the need for fine-tuning and precisely balancing exploration and exploitation in real-time.

	\begin{table}[!htb]
		
		\centering
		\begin{tabular}{@{}>{\raggedright\arraybackslash}p{3.2cm} 
				>{\centering\arraybackslash}p{2.3cm} 
				>{\centering\arraybackslash}p{2.3cm} 
				>{\centering\arraybackslash}p{2.3cm} 
				>{\raggedright\arraybackslash}p{3.1cm}@{}}
			\toprule
			Algorithm         & Linear Modeling    & Local History Modeling & Attention Mechanism  & \(k\) Selection Method          \\ \midrule
			UCB               & No                 & No                     & No                   & N/A                            \\
			KL-UCB            & No                 & No                     & No                   & N/A                            \\
			$k$-NN UCB        & No                 & Yes                    & No                   & Function optimization          \\
			$k$-NN KL-UCB     & No                 & Yes                    & No                   & Function optimization          \\
			LinThompson       & Yes                & No                     & No                   & N/A                            \\
			LinThompsonUCB    & Yes                & No                     & No                   & N/A                            \\
			LinUCB            & Yes                & No                     & No                   & N/A                            \\
			\textbf{LNUCB-TA} & \textbf{Yes}       & \textbf{Yes}           & \textbf{Yes}         & \textbf{Variance guided, nonparametric} \\
			\bottomrule
		\end{tabular}
		\caption{Key attributes in our approach compared to existing MAB algorithms. ``Yes" indicates the presence of the feature,``No" indicates the absence of the feature, and ``N/A" indicates not applicable.}    
        \label{comparison-table-CONTRIBUTION}
	\end{table}
	\paragraph{Motivating examples.}
	One application of the proposed hybrid model is in online advertisement recommendation, aiming to maximize user engagement through demographics, browsing history, and time-specific data \citep{zeng2016online}. The linear component captures broad trends, such as higher click-through rates for fashion advertisements among users aged 18 to 25, while the adaptive {$k$}-NN component refines this by recognizing local patterns. For instance, users within the 18 to 25 age group who frequently visit sports websites might prefer sports equipment advertisements. Furthermore, the novel exploration mechanism dynamically balances exploring new advertisement types and exploiting known preferences, thus optimizing real-time recommendations by leveraging both global trends and individual user behaviors.
	
	Another application is in the exploration of partially observed social networks to maximize node discovery within a set query budget \citep{madhawa2019multi}, where our proposed hybrid model proves beneficial. The linear model identifies nodes with high-degree centrality as valuable targets based on their potential to connect to many others. The adaptive {$k$}-NN model enhances this strategy by pinpointing densely connected sub-communities within these high-centrality nodes, likely revealing new nodes when queried. Meanwhile, the attention mechanism dynamically shifts the exploration and exploitation based on the real-time performance of each node, enhancing the efficiency of network exploration by focusing on nodes that show promising connectivity trends while still exploring lesser-known parts of the network.
	
	\paragraph{Organization.} The rest of the paper is structured as follows. Section \ref{sec: hybrid} covers the rigorous mathematical setup of the problem. Section \ref{sec:methodology} presents the LNUCB-TA algorithm. 
	The theoretical analysis of the algorithm is presented in Section \ref{sec:theory}. Section \ref{sec:result} provides the experimental results. Conclusions are discussed in Section \ref{sec:conclusion}. Detailed proofs of the theoretical results, additional findings, limitations, future research directions, and implementation guidelines are included in the Appendix.

	\section{Hybrid Contextual MAB Learning}
	\label{sec: hybrid}
	
	\paragraph{Problem Definition.}
	
	We consider a hybrid contextual MAB problem within a metric space \((\mathcal{X} \times \mathcal{Z}, \rho)\), where \(\mathcal{X} \times \mathcal{Z}\) represents the joint space of context features and reward history. Time is indexed discretely as \( t = 1, 2, \ldots, T \), where \( T \) is the total number of time steps. Each context \(x_t \in \mathcal{X}\) at time \(t\) corresponds to a set of possible actions, or ``arms," indexed by \(a\) within the set \(\mathcal{A} = \{1, \ldots, A\}\), where \(A\) is the total number of arms. The reward corresponding to each arm \(a\) for a given context \(x_t\) at time \(t\) is denoted as a random variable \(Y_t^a\), constrained within the interval \([-1,1]\). The vector \( Y_t = (Y_t^a)_{a \in \mathcal{A}} \in \mathbb{R}^A \) comprises the stochastic rewards for all arms at time \( t \), and the random variable \( Y_t^a \) is defined conditionally on the context and the history of previous rewards. Upon observation, the realized reward for arm \(a\) is given by:
	\begin{equation}
	\hat{Y}_t^a = o_t^a(x_t, z_t) + \xi_t^a,
	\end{equation}
	where \( \xi_t^a \) is the noise term, capturing stochastic errors not explained by the model predictions for arm \(a\). Here, the expected reward for arm \(a\) at time \(t\) is given by the function \(o^a_t: \mathcal{X} \times \mathcal{Z} \rightarrow [-1,1]\), defined as:
	\begin{equation}
		o^a_t(x_t^a, z_t^a) = \mathbb{E}[Y_t^a \mid X_t = x_t, Z_t = z_t] = l^a_t(x_t^a) + f_{k,t}^a(x_t^a, z_t^a) = \mu_t^a)^\top x_t^a + \text{k-NN}_{k,t}^a(x_t^a, z_t^a),\label{eq:reward}
	\end{equation}
	where \(\mathbb{E}\) denotes the expectation, and \(z_t^a = \{\hat{Y}_s^a : s < t, a \in \mathcal{A}\}\) represents the observed historical rewards for arm \(a\) up to time \(t\) with \(z_t \in \mathcal{Z}\). The feature vector \(X_t\) is drawn independently and identically distributed (i.i.d.) from a fixed marginal distribution \(\mathbb{D}\) over \(\mathcal{X}\). The linear model's prediction for arm \( a \) given the context \( x_t^a \), which represents the feature vector for arm \( a \) at time \( t \), is given by \(l_t^a(x_t^a) = (\mu_t^a)^\top x_t^a\), where \( \mu_t^a \) is the parameter vector associated with arm \( a \) at time \( t \).
    On the other hand, the estimation of the \(k\)-NN model for each arm is based on the corresponding historical observed rewards for the selected \(k\) nearest neighbors up to time \(t\). At each time step \(t\), the number of neighbors \(k_t^a\) is dynamically determined based on the variance of the reward history for arm \(a\), ensuring that the model adapts to changes in the reward distribution over time. The \(k\)-NN estimation is formally defined as:  

\begin{equation}
f_{k,t}^a(x_t^a, z_t^a) = \frac{1}{k_t^a} \sum_{s \in N_{k_t^a}(x_t^a)} \hat{Y}_s^a,
\end{equation}

	where \( \hat{Y}_s^a \) represents the observed reward for arm \(a\) at time step \(s\) (with \(s < t\)). The set of neighbors \( N_{k_t^a}(x_t^a) \) consists of the indices of the \(k_t^a\)-nearest neighbors to \(x_t^a\), selected based on the Euclidean distance within the contextual feature space. Consequently, \( f_{k,t}^a \) relies solely on the observed rewards from \(z_t^a\) for the closest neighbors in terms of context similarity.  Furthermore, for any context \(x_t^a \in \mathcal{X}\) and a radius \(r > 0\), \(\text{BALL}_t^a(x_t^a, r)\) denotes the open metric ball centered at \(x_t^a\) with radius \(r\) for arm \(a\). This metric ball is pivotal for analyzing distances and neighborhood relations within the joint space \(\mathcal{X} \times \mathcal{Z}\).

	\paragraph{Decision Policy.}
	
	The decision-making process within the hybrid contextual MAB framework is guided by a policy \(\pi = \{\pi_t\}_{t \in [T]}\), where each policy function \(\pi_t : \mathcal{X} \times \mathcal{Z} \to [A]\) maps the observed context and reward history to an arm. This mapping is based on the integration of linear estimation and nonlinear estimation utilizing the historical data \(\mathcal{H}_{t-1} = \{(X_s, \pi_s, Y_s^{\pi_s})\}_{s \in [t-1]}\), which consists of previously observed contexts, the arms chosen, and the corresponding rewards.
	
	\paragraph{Exploration-Exploitation Trade-Off.}
	
	In our problem, the exploration-exploitation \sloppy{trade-off}
	is managed through a dynamic, attention-based exploration factor. This approach adapts the exploration parameter \(\alpha\) in real-time based on both global performance (\(g\)) and specific reward patterns of individual arms (\(n_t^a\)), ensuring a more balanced and effective strategy. The exploration parameter \(\alpha\) is updated dynamically according to:
	\begin{equation}
		\alpha_{N_t^a} = \frac{\alpha_0}{N_t^a + 1} \cdot \left( \kappa g + (1 - \kappa) n_t^a \right),
	\end{equation}
    where the weight factor \(\kappa\) adjusts the balance between global and local rewards, with lower values enhancing adaptability and higher values favoring stability, \(n_t^a = \frac{1}{N_t^a} \sum_{\hat{Y}_s^a \in z_t^a} \hat{Y}_s^a = \frac{1}{N_t^a} \sum_{s=1}^{t-1} \hat{Y}_s^a\)
	represents the average reward history of arm \(a\) up to time \(t\) (reward patterns of an individual arm), with \(N_t^a = |\hat{Y}_{1:t-1}^a|\) as the number of pulls of arm \(a\) up to time \(t\). If \(N_t^a = 0\), \(n_t^a\) is set to zero.
	
	\paragraph{Objective.}
	
	The primary aim is to maximize the cumulative reward over \(T\) time steps, represented by \(\sum_{t \in [T]} Y_t^{\pi_t}\), and to minimize the regret relative to an oracle policy \(\pi^* = \{\pi_t^*\}_{t \in [T]}\), where \(\pi_t^* = \arg\max_{a \in [A]} o^a_t(x_t, z_t)\). In LNUCB-TA, the optimal decision \((\pi_t^{a})^*\) through the optimal context \((x_t^{a})^*\) for each arm would be the decision that maximizes the expected combined reward based on the linear model predictions and the adjustments made by the \(k\)-NN model, using the best available historical data up to time step \(t\) defined as:
	\begin{equation}
		(x_t^a)^* \in \arg\max_{x \in D} \left( (\mu^a)^* \cdot (x_t^a) + f_{k,t}^a(x_t^a, z_t^a) \right),
	\end{equation}
	where \((\pi_t^a)^*\) refers to the best reward obtained for arm \(a\) based on its history over \(t\) steps, which leads to the theoretical optimal action \(\pi_t^*\), and \(D\) represents the decision space. Although we compute an optimal action for each arm, the model ultimately selects only one arm to play per time step, choosing the one with the highest expected reward. \((\mu^a)^*\) is the best estimate of the parameter vector across arm \(a\), assuming an oracle setting, or the true underlying model known retrospectively.
	
	\paragraph{Regret Analysis.}
	
	The regret, \(R_T(\pi)\), is a measure of the performance difference and is defined as:
	\begin{equation}
		R_T(\pi) = \sum_{t \in [T]} (Y_t^{\pi_t^*} - Y_t^{\pi_t}).
	\end{equation}
	In our proposed model, for a single arm \(a\), the regret at time \(t\) can be defined as:
	\begin{equation}
		\text{regret}_t^a = \Delta_{t}^a \left( g_t^a\left( (x_t^a)^*, (z_t^a)^* \right) - o_t^a\left( x_t^a, z_t^a \right) \right),\label{eq:regret}
	\end{equation}
	where \(g_t^a\left( (x_t^a)^*, (z_t^a)^* \right)\) is the optimal expected reward for arm \(a\) at the optimal context \((x_t^a)^*\), which is the feature vector that would yield the highest reward for arm \(a\), leading to optimal \((z_t^a)^*\), and \(\Delta_{t}^a\) is the indicator function that equals 1 if arm \(a\) is selected at time \(t\) and 0 otherwise. The function \(o^a_t(x_t^a, z_t^a)\) represents the expected reward under the decision made by the policy \(\pi_t^a\) at context \(x_t^a\) with reward history of \(z_t^a\). As a result, the total cumulative regret for LNUCB-TA over a time horizon \(T\) across all arms is calculated as:
	\begin{equation}
		\begin{split}
			R_T = & \sum_{a=1}^A \sum_{t=0}^T \Delta_{t}^a \left( g^a\left( (x_t^a)^*, (z_t^a)^* \right) - o_t^a\left( x_t^a, z_t^a \right) \right) \\
			= & \sum_{a=1}^A \sum_{t=0}^T \Delta_{t}^a \left( l_t^a\left( (x_t^a)^* \right) + f_{k,t}^a\left( (x_t^a)^*, (z_t^a)^* \right) - \left( l_t^a(x_t^a) + f_{k,t}^a\left( x_t^a, z_t^a \right) \right) \right) \\
			= & \sum_{a=1}^A \sum_{t=0}^T \Delta_{t}^a \left( (\mu_t^a)^* \cdot (x_t^a)^* + \text{k-NN}_{k,t}^a\left( (x_t^a)^*, (z_t^a)^* \right) - \left( \mu_t^a)^\top x_t^a + \text{k-NN}_{k,t}^a\left( x_t^a, z_t^a \right) \right) \right).\label{eq:regret-total}
		\end{split}
	\end{equation}
	
	\section{Methodology}
	\label{sec:methodology}
	
	 In the following sections, we present an overview of the proposed approach, outlining its core components and how they contribute to its improved performance. The first part of this methodology section describes the overall conceptual framework, followed by a detailed explanation of the algorithmic structure and implementation.
	
	\subsection{Overall Concept}
	
	\paragraph{Intuition.} 
	We propose the LNUCB-TA model, which introduces two significant innovations to previously proposed contextual UCB algorithms. Both of these advancements enhance the adaptability and accuracy in dynamic environments. The proposed method, shown in Algorithm~\ref{alg:lnucb_ta}, is initiated using the structural framework of the LinUCB algorithm, which employs a linear model to estimate the rewards for each arm \(a\) based on contextual features indicated as \( l_t^a = \left( x_t^a \right)^\top \mu_t^a\). This basic linear framework is then augmented using a nonlinear component through the use of the {$k$}-Nearest Neighbors method. This enhancement integrates insights from the history of both the reward and context, and effectively captures the recent profile of the features (Algorithm~\ref{alg:k-NN}).
	
	In addition to refining reward estimations, our approach introduces an attention-based exploration factor, \(\alpha_{N_t^a}\), which tunes the exploration-exploitation balance dynamically (Algorithm~\ref{alg:attention}). This defines the dynamic Upper Confidence Bound, which balances exploration and exploitation to select the optimal arm, extending the LinUCB framework proposed by \citep{li2010contextual}, which uses a fixed exploration parameter, by introducing a dynamically adjusted exploration factor.

    \begin{equation}
        UCB_t^a = \left(l_t^a + \text{\{$k$-NN\} score} \right) + (\alpha_{N_t^a}) \cdot \sqrt{(x_t^a)^\top (\Sigma_t^a)^{-1} x_t^a}
    \end{equation}
	
	\begin{algorithm}
		\caption{LNUCB-TA}
		\begin{algorithmic}[1]
			\State \textbf{Input:}  $\lambda,\beta,\alpha_0, \kappa$ \Comment{Model parameters}
			\For{$t = 0, 1, 2, \ldots$}
			\For{each arm $a$ in $A$}
			\State Compute $l_t^a = (x_t^a)^\top \mu_t^a$ \Comment{Linear estimation}
			\State Compute {$k$-NN score (reward adjustment)} \Comment{Nonlinear estimation}
			\State Compute $UCB_t^a$ based on attention-based exploration rate \Comment{Dynamic UCB}
			\EndFor
			\State Select arm $a_t = \arg\max_{a \in A} \left( UCB_t^a\right)$
			\State Update $\text{BALL}_{t+1}^a$ and model parameters \Comment{Uncertainty region}
			\EndFor
		\end{algorithmic}
		\label{alg:lnucb_ta}
	\end{algorithm}
	
	\paragraph{Method.} 
	The LNUCB-TA model, shown in Algorithm~\ref{alg:lnucb_ta}, not only maintains the structure of the original LinUCB framework but also seamlessly integrates adaptive nonlinear adjustments and real-time refinements in confidence bounds and exploration rates. These enhance the model's adaptability and accuracy in complex environments. Through this careful augmentation, we extend the LinUCB's capability while preserving its theoretical underpinnings, ensuring that our contributions are both innovative and robustly grounded in established methodologies. In the following section, the two novel components are discussed in more detail.
	
	\subsection{Nonlinear Estimation Using Feature and Reward History}
	
	\paragraph{Intuition.} 
	The adaptive {$k$}-NN ensures that the model adjusts its reliance on the reward history of each arm based on the stability of the rewards. It seamlessly integrates more insights from {$k$}-NN as additional data becomes available and defaults to a more conservative approach when data is sparse. This unique method effectively captures local patterns with improved time efficiency, without the need for extensive function optimization, thereby enhancing adaptability and responsiveness in dynamic environments.
	
	\paragraph{Method.} 
	The adaptive {$k$}-NN strategy employed in the model, detailed in Algorithm~\ref{alg:k-NN}, takes both the reward history and the feature vector of each arm as inputs. This method is applied conditionally, specifically when the length of the feature vector \(x_t^a\) (where \(x_t^a\) represents the contextual features of arm \(a\) at time \(t\)) is greater than or equal to the number of neighbors \(k_t^a\) (where \(k_t^a\) is the dynamically determined number of nearest neighbors for arm \(a\) at time \(t\)). This ensures sufficient historical data is available for accurate neighbor selection and reward estimation.
	
	\begin{algorithm}
		\caption{Adaptive {$k$}-NN Integration for LNUCB-TA}
		\begin{algorithmic}[1]
			\State \textbf{Input:} Decision space \(D\), Historical data \(\mathcal{H}\), \(\theta_{\text{min}}\) and \(\theta_{\text{max}}\) to determine the number of neighbors
			\State Observe context \(X_t\), Reward history \(Z_t\)
			\For{\textbf{each} arm \(a\) in \(\mathcal{A}\) at time \(t\)}
			\State Compute variance of rewards \( \text{Var}(z_t^a) \) \Comment{Reward variance}
			\State \(k_t^a = \theta_{\text{min}} + (\theta_{\text{max}} - \theta_{\text{min}}) \times \text{Var}(z_t^a)\)
			\If{\(\text{len}(x_t^a) \geq k_t^a\)}
			\State \( f_{k,t}^a(x_t^a, z_t^a) = \text{{$k$}-NN}_{k,t}^a(x_t^a, z_t^a) \) \Comment{{$k$}-NN-score}
			\State Estimated reward = \( l^a_t(x_t^a) + f_{k,t}^a(x_t^a, z_t^a) \) \Comment{Reward estimation}
			\State Model update = \( \text{UCB}_t^a \)
			\State Select arm \(a\) with the highest updated model prediction
			\EndIf
			\EndFor
		\end{algorithmic}
		\label{alg:k-NN}
	\end{algorithm}
	
	In Algorithm~\ref{alg:k-NN}, the variance in rewards for each arm at time \(t\), \(\text{Var}(z_t^a)\), drives the adaptive selection of \(k\), which influences the depth of historical data employed for the {$k$}-NN based prediction. The \(k\) value dynamically adjusts between predefined minimum (\(\theta_{\text{min}}\)) and maximum (\(\theta_{\text{max}}\)) thresholds. The selection of \(\theta_{\min}\) and \(\theta_{\max}\) is determined through hyperparameter tuning as they define the range for \(k\), based on the observed variability of rewards, where:
	
	\begin{itemize}
		\item \textbf{Low Variance:} Indicates stable reward patterns, suggesting that fewer historical data points are sufficient for accurate predictions. This stability allows the model to maintain a smaller \(k\), closer to the minimum threshold, optimizing computational efficiency while maintaining predictive accuracy.
		\item \textbf{High Variance:} Reflects irregular or unpredictable reward patterns, necessitating a larger \(k\) to incorporate a broader historical context. This expanded view helps to mitigate the impact of variability, enhancing the robustness of reward predictions.
	\end{itemize}
	
	Furthermore, unlike existing nonlinear approaches that use a static {$k$} or involve searching over the preceding time steps \(k \in [1, t-1]\) \citep{park2014greedy, reeve2018k}, our proposed model adopts a data-driven approach for selecting \(k\). The algorithm achieves a time complexity of \(O(t)\), which can reach \(O(1)\) per update in the optimal case, significantly decreasing time complexity compared to the function optimization techniques used in {$k$}-NN UCB and {$k$}-NN KL-UCB.
	
	\subsection{Temporal Attention}
	
	\paragraph{Intuition.} 
	Our model replaces static exploration parameters with an attention-based mechanism, which allows for dynamic adjustment of exploration efforts based on time-dependent changes (temporal) and distinct reward patterns across different arms or contexts (spatial). The proposed method analyzes global performance across all arms, specific reward patterns of individual arms, and the frequency of arm selections, dynamically adjusting \(\alpha\) for each arm at each time step. This innovation leads to consistent results, independent of the initial choice of the exploration rate.
	
	\begin{algorithm}
		\caption{Temporal Attention-Based Exploration Rate for LNUCB-TA}
		\begin{algorithmic}[1]
			\State \textbf{Input:} \(\alpha_0\), \(N_t^a\) (number of times arm \(a\) played up to \(t\)), \(g\) as global average of rewards, \(n_t^a\) as mean average of each arm, \(\kappa\) as weight factor
			\For{\textbf{each} arm \(a\) in \(\mathcal A\) at time \(t\)}
			\State \(n_t^a = \frac{1}{N_t^a} \sum_{\hat{Y}_s^a \in z_t^a} \hat{Y}_s^a\)
			\Comment Local attention for arm \(a\)
			\State \( \alpha_{N_t^a} = \frac{\alpha_0}{(N_t^a + 1)} \cdot (\kappa g + (1 - \kappa) n_t^a) \)
			\Comment Attention-based exploration factor
			\State Update \(UCB_t^a\)
			\EndFor
		\end{algorithmic}
		\label{alg:attention}
	\end{algorithm}
	
	\paragraph{Method.} 
	As shown in Algorithm~\ref{alg:attention}, the attention-based exploration rate \(\alpha_{N_t^a}\) dynamically decreases as an arm is played more frequently, reflecting a transition from exploration to exploitation. This adaptive mechanism reduces uncertainty about arm performance over time. Simultaneously, the local reward \(n_t^a\) plays a crucial role in refining the exploration factor—higher local rewards for an arm increase its likelihood of further exploration, ensuring that promising arms are not prematurely discarded. By incorporating both global (\(g\)) and local (\(n_t^a\)) attention components, the algorithm effectively balances exploration and exploitation without requiring manual tuning of exploration parameters. The global attention component ensures broad coverage across all arms, preventing excessive focus on a few high-performing ones, while the local attention component allows fine-grained adaptation to individual arm dynamics. This results in a more consistent and adaptive exploration strategy compared to traditional MAB models, which typically rely on static or heuristically adjusted exploration rates.
	
	\section{Theoretical Analysis}
	\label{sec:theory}
	
	\begin{theorem}[Regret Bound]
		\label{thm:regret}
		Suppose the noise \( \left|\xi_{t}^a\right| \) is bounded by \( \sigma \) (i.e., \( \left|\xi_{t}^a\right| \leq \sigma \)), the true parameter vector \( (\mu^a)^* \) has a norm bounded by \( W \) (i.e., \( \left\|(\mu^a)^*\right\| \leq W \)), and the context vectors \( x \) are bounded such that \( \left\| x \right\| \leq B \) for all \( x \in D \), and let \( \lambda = \frac{\sigma^2}{W^2} \). Then parameter \( \beta_t^a \) can be defined as:
		\begin{equation}
			\beta_t^a := \sigma^2 \left( 2 + 4d \log \left( 1 + \frac{TB^2 W^2}{d} + \frac{\sum_{a=1}^A T^a (u_{t,k}^a)^2}{d} \right) + 8 \log \left( \frac{4}{\delta} \right) \right),
		\end{equation}
		with probability greater than \( 1 - \delta \), for all \( t \geq 0 \),
		\begin{equation}
			R_T \leq b \sigma \sqrt{T \left( d \log \left( 1 + \frac{TB^2 W^2}{d \sigma^2} + \frac{\sum_{a=1}^A T^a (u_{t,k}^a)^2}{d \sigma^2} \right) + \log \left( \frac{4}{\delta} \right) \right)},
		\end{equation}
		where \( \sigma^2 \) represents the total variance accounting for both the linear component and the additional variance from the \( k \)-NN model, \( \delta \) is the probability with which the confidence bounds are held, \( b \) is an absolute constant, and \( \sum_{a=1}^A T^a (u_{t,k}^a)^2 \) represents the sum of the squared uncertainties for each arm \( a \), capturing the influence of \( k \)-NN's neighborhood-based uncertainty for each specific arm. This sum is scaled by the number of times each arm \( a \) is played \( T^a \), where \( \sum_{a=1}^A T^a \leq T \) as not all arms may apply the \( k \)-NN adjustment at every time step. This sum represents an upper bound, capturing the maximum possible contribution from the \( k \)-NN component.
		
		Given these conditions, the simplified regret bound for LNUCB-TA is \( R_T = \mathcal{O}(\sqrt{dT \log T}) \), which, by absorbing logarithmic factors into \( \tilde{\mathcal{O}} \), we can state:
		\begin{equation}
			R_T = \tilde{\mathcal{O}}(\sqrt{dT}).
		\end{equation}
	\end{theorem}
	
	This bound demonstrates that LNUCB-TA achieves sub-linear regret, highlighting its diminishing regret growth rate over time, contrasting with linear regret, where regret scales linearly with time steps. To prove this theorem, we need to establish the following two propositions:
	
	\begin{proposition}[Uniform Confidence Bound]
		\label{prop:ucb}
		Let \( \delta > 0 \). We have
		\begin{equation}
			\Pr\left( \forall t, (\mu^a)^* \in \text{BALL}_{t}^a \right) \geq 1 - \delta.
		\end{equation}
	\end{proposition}
	
	The second key proposition in analyzing LNUCB-TA involves demonstrating that, provided the aforementioned high-probability event occurs, the growth of the regret can be effectively controlled. Let us define the instantaneous regret as:
	\begin{equation}
		\text{regret}_t^a = (\mu^a)^* \cdot (x_t^a)^* + \text{{$k$}-NN}_{k,t}^a\left( (x_t^a)^*, (z_t^a)^* \right) - \left( (\mu^a)^* \cdot x_t^a + \text{{$k$}-NN}_{k,t}^a(x_t^a, z_t^a) \right).
	\end{equation}
	The following proposition provides an upper bound on the sum of the squares of the instantaneous regret.
	
	\begin{proposition}[Sum of Squares Regret Bound]
		\label{prop:sum-regret}
		Suppose \( \|x\| \leq B \) for all \( x \in D \), and assume \( (\mu^a)^* \in \text{BALL}_t^a \) for all \( t \). Then, the sum of the squares of instantaneous regret for each arm \( a \) over time is bounded as:
		\begin{equation}
			\sum_{t=0}^{T-1} (\text{regret}_t^a)^2 \leq 8 \beta_t^a d \log \left( 1 + \frac{TB^2}{d\lambda} + \frac{\sum_{a=1}^A T^a (u_{t,k}^a)^2}{d\lambda} \right).
		\end{equation}
		The cumulative squared regret bound is given by:
		\begin{equation}
			\sum_{a=1}^A \sum_{t=0}^{T-1} \text{regret}_t^a \leq \sqrt{T \sum_{t=0}^{T-1} (\text{regret}_t^a)^2}
			\leq \sqrt{8T\beta_t^a d \log \left( 1 + \frac{TB^2}{d\lambda} + \frac{\sum_{a=1}^A T^a (u_{t,k}^a)^2}{d\lambda} \right)}.
		\end{equation}
	\end{proposition}
	
	\begin{theorem}[Temporal Exploration-Exploitation Balance]
		\label{thm:exploration}
		Given a set of arms \( \{1, 2, \dots, A\} \) in a MAB problem, where each arm \( a \) has a set of observed rewards denoted by \( Y_t = (Y_t^a)_{a \in \mathcal{A}} \) in \( \mathbb{R}^A \), and \( N_t^a \) is the number of times arm \( a \) has been selected up to time \( t \). An attention mechanism can be designed, which dynamically updates the exploration parameter \( \alpha \) according to the formula:
		\begin{equation}
			\alpha_{N_t^a} = \frac{\alpha_0}{N_t^a + 1} \cdot \left( \kappa g + (1 - \kappa) n_t^a \right),
		\end{equation}
		where \( g \) represents the global attention derived from the average rewards across all arms, \( n_t^a \) represents local attention derived from the average reward of arm \( a \) at time \( t \), and \( \kappa \) is a weighting factor that balances global and local attention components.
	\end{theorem}
	
	\section{Results}
	\label{sec:result}
	
    Our comprehensive evaluation establishes the superiority of LNUCB-TA through three key analyses: (1) benchmark comparisons against eight state-ppf-the-art (SOTA) bandit algorithms, covering linear, nonlinear, kernel-based, and neural-based models, across four standard datasets; (2) validation in news recommendation scenarios, where LNUCB-TA is tested against more than ten MAB models on a large-scale dataset; and (3) component-level ablation studies, which break down and assess the contributions of LNUCB-TA’s architectural innovations. We first demonstrate its effectiveness across diverse real-world domains, then evaluate its performance in high-dimensional recommendation systems, and finally analyze the impact of its design choices through controlled experiments. Additional validations are provided in Appendix \ref{sec:more results}, covering parameter analysis (Figures \ref{fig:cumulative}-\ref{fig:mean}, Table \ref{table:comparison-table}), model improvements (Figure \ref{fig:enhanced}, Table \ref{table:comparison-table-enhanced}), error bars (Figure \ref{fig:errorbars}), and generalization (Figures \ref{fig:network_dataset}-\ref{fig: runtime scalability}, Table \ref{table:article_matching}).
 
	\subsection{Benchmark Performance}
	\label{subsec:benchmarks}
	
    This section presents an empirical evaluation of our algorithm on several public benchmark datasets, including Adult, Magic Telescope, and Mushroom from the UCI repository \citep{asuncion2007uci}, as well as MNIST \citep{yann2010mnist}. Table ~\ref{table:regret_comparison_final} reports the total regret and standard deviation across datasets, averaged over 20 runs.

	\begin{table}[!htb]
		\centering
		\small
		\begin{tabular}{@{}>{\raggedright\arraybackslash}p{2.5cm} *{4}{>{\centering\arraybackslash}p{2cm}}@{}}
			\toprule
			\textbf{Model} & \textbf{Adult} & \textbf{Magic} & \textbf{MNIST} & \textbf{Mushroom} \\
			\midrule
			Linear UCB & $2097.5 \pm 50.3$ (2.40\%) & $2604.4 \pm 34.6$ (1.33\%) & $2544.0 \pm 235.4$ (9.25\%) & $562.7 \pm 23.1$ (4.11\%) \\
			Linear TS & $2154.7 \pm 40.5$ (1.88\%) & $2700.5 \pm 46.7$ (1.73\%) & $2781.4 \pm 338.3$ (12.16\%) & $643.3 \pm 30.4$ (4.72\%) \\
			Kernel UCB & $2080.1 \pm 44.8$ (2.15\%) & $2406.5 \pm 79.4$ (3.30\%) & $3595.8 \pm 580.1$ (16.13\%) & $199.0 \pm 41.0$ (20.60\%) \\
			Kernel TS & $2111.5 \pm 87.4$ (4.14\%) & $2442.6 \pm 64.5$ (2.64\%) & $3406.0 \pm 411.7$ (12.09\%) & $291.2 \pm 40.0$ (13.74\%) \\
			BootstrapNN & $2097.3 \pm 39.3$ (1.87\%) & $2269.4 \pm 27.9$ (1.23\%) & $1765.6 \pm 321.1$ (18.19\%) & $132.3 \pm 8.6$ (6.50\%) \\
			$\epsilon$-greedy & $2328.5 \pm 50.4$ (2.16\%) & $2381.8 \pm 37.3$ (1.57\%) & $1893.2 \pm 93.7$ (4.95\%) & $323.2 \pm 32.5$ (10.06\%) \\
			NeuralUCB & $2061.8 \pm 42.8$ (2.08\%) & $2033.0 \pm 48.6$ (2.39\%) & $2071.6 \pm 922.2$ (44.49\%) & $160.4 \pm 95.3$ (59.41\%) \\
			NeuralTS & $2092.5 \pm 48.0$ (2.29\%) & $2037.4 \pm 61.3$ (3.01\%) & $1583.4 \pm 198.5$ (12.53\%) & $115.0 \pm 35.8$ (31.13\%) \\
			\textbf{LNUCB-TA (Ours)} & \textbf{1673.1} $\pm 12.07$ (0.72\%) & \textbf{1931.6} $\pm 31.22$ (1.62\%) & \textbf{1561.6} $\pm 42.09$ (2.69\%) & \textbf{19.85} $\pm 1.98$ (9.97\%) \\
			\bottomrule
		\end{tabular}
		\caption{Comparison of \textbf{LNUCB-TA} vs different types of bandit models based on regret (Mean ± Standard Deviation and relative Std/Mean percentage over 20 runs.}
		\label{table:regret_comparison_final}
	\end{table}
    
    Following \citep{zhang2021neural}, the datasets were randomly shuffled, and features were normalized such that their $\ell_2$-norm equals unity. For NeuralTS, a one-hidden-layer neural network with 100 neurons has been applied. This is done by using gradient descent for posterior updates with 100 iterations and a learning rate of 0.001. The BootstrapNN method has utilized an ensemble of 10 identical networks, where each data point is included for training with a probability of 0.8 at each round ($p = 10$, $q = 0.8$), following the original work of \citep{schwenk2000boosting}. For $\epsilon$-Greedy, the exploration parameter $\epsilon$ has been tuned via grid search over $\{0.01, 0.05, 0.1\}$. In Linear and Kernel UCB / Thompson Sampling, the regularization parameter has been set to $\lambda = 1$ as in \citep{agrawal2013thompson, chowdhury2017kernelized} and optimized the exploration parameter $\nu$ using a grid search over $\{1, 0.1, 0.01\}$. For neural UCB and neural Thompson sampling, grid searches have been performed on $\lambda \in \{1, 10^{-1}, 10^{-2}, 10^{-3}\}$ and $\nu \in \{10^{-1}, 10^{-2}, 10^{-3}, 10^{-4}, 10^{-5}\}$ to determine the best-performing configuration.
	
	As shown in Table~\ref{table:regret_comparison_final}, our proposed LNUCB-TA model outperforms the baseline models across multiple datasets. Notably, in the Adult dataset, LNUCB-TA achieves a regret of $1673.1 \pm 12.07$, outperforming the next best-performing model, Kernel UCB , which has a regret of $2080.1 \pm 44.8$, showcasing a significant improvement. In the Magic dataset, LNUCB-TA again leads with a regret of $1931.6 \pm 31.22$, surpassing the NeuralUCB model, which has a regret of $2033.0 \pm 48.6$. In the MNIST dataset, LNUCB-TA holds a slight edge over the NeuralTS model, which shows a regret of $1583.4 \pm 198.5$, achieving $1561.6 \pm 42.09$. Finally, in the Mushroom dataset, LNUCB-TA’s remarkable performance of $19.85 \pm 1.98$ shows a substantial improvement compared to other models, further emphasizing the strength of our model in minimizing total regret across diverse settings.

	\subsection{News Recommendation Dataset}
	\label{subsec:news_case_study}
	
	This section evaluates LNUCB-TA on a benchmark news recommendation dataset with 10,000 entries, each containing 102 features. The first feature indicates one of ten news articles, the second represents user engagement (click/no click), and the remaining features provide contextual information \citep{li2010contextual, li2011unbiased}. Both the estimated reward and its variability serve as critical metrics in our analysis.

	\begin{figure}[ht]
		\begin{center}
			\fbox{\includegraphics[width=0.98\textwidth]{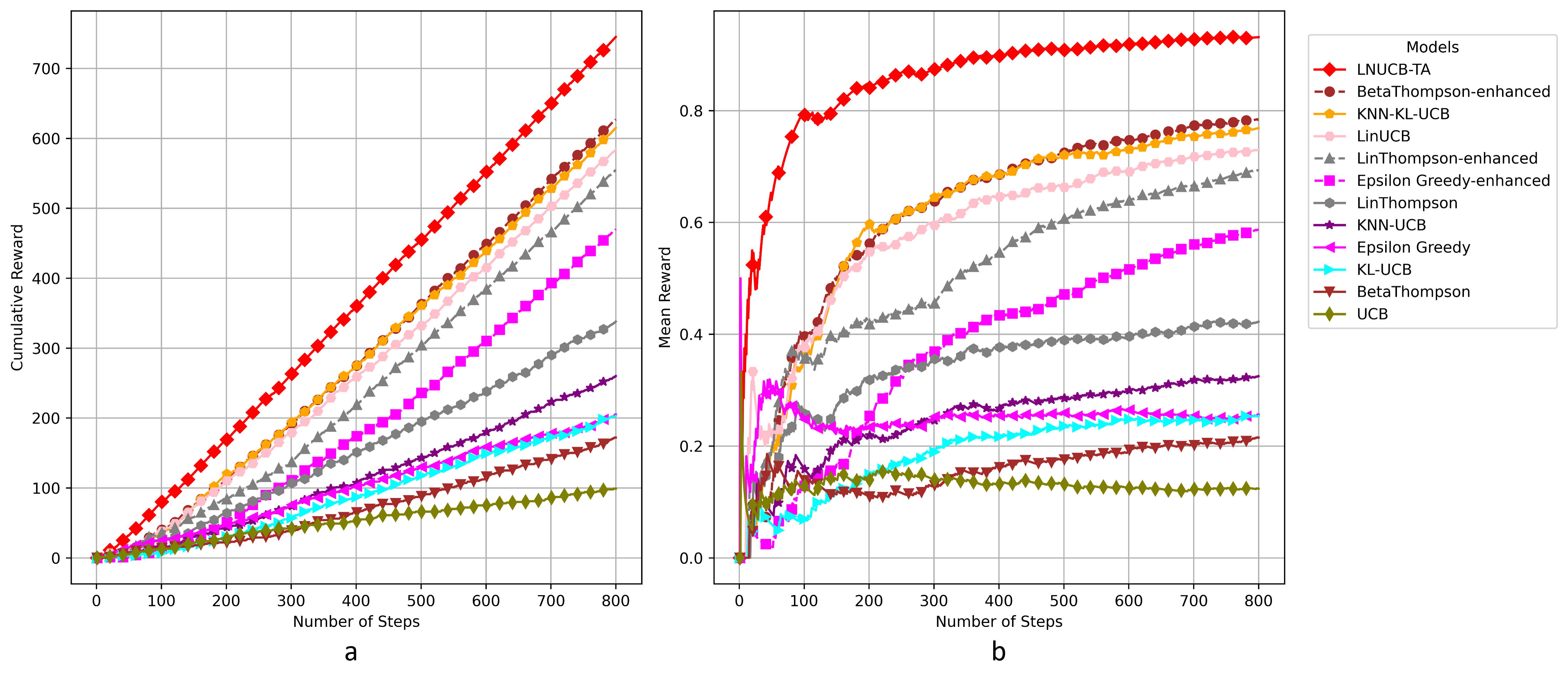}} 
		\end{center}
		\vspace{1em} 
		\caption{(a) Cumulative rewards over 800 steps for \textbf{LNUCB-TA} and other models, sowing LNUCB-TA's superior performance. (b) Mean rewards per time step, highlighting LNUCB-TA's rapid convergence and consistent high performance.}
		\label{fig:my_label}
	\end{figure}

    	\begin{table}[ht]
		\begin{center}
			\begin{tabular}{@{}>{\raggedright\arraybackslash}p{3.2cm} 
					>{\centering\arraybackslash}p{2cm} 
					>{\centering\arraybackslash}p{2cm} 
					>{\centering\arraybackslash}p{1.8cm} 
					>{\centering\arraybackslash}p{1.8cm} 
					>{\centering\arraybackslash}p{1.8cm}@{}}
				\toprule
				Model            & Exploration Rate ($\alpha/\rho$) & Cumulative Reward & Mean Reward & Run Time (s) & Std Dev of Mean Reward \\ \midrule
				(Lin+$k$-NN)-UCB   & 0.1                              & 662               & 0.83        & 715.02       & 0.35                  \\
				(Lin+$k$-NN)-UCB   & 1                                & 617               & 0.77        & 733.72       & 0.35                  \\
				(Lin+$k$-NN)-UCB   & 10                               & 160               & 0.20        & 758.82       & 0.35                  \\
				LinUCB           & 0.1                              & 567               & 0.71        & 8.09         & 0.30                  \\
				LinUCB           & 1                                & 424               & 0.53        & 8.73         & 0.30                  \\
				LinUCB           & 10                               & 98                & 0.12        & 5.97         & 0.30                  \\
				$k$-NN UCB         & 0.1                              & 195               & 0.24        & 459.71       & 0.05                  \\
				$k$-NN UCB         & 1                                & 192               & 0.24        & 434.08       & 0.05                  \\
				$k$-NN UCB         & 10                               & 260               & 0.33        & 457.07       & 0.05                  \\
				\textbf{LNUCB-TA}         & 0.1                              & \textbf{741}                & \textbf{0.93}        & 324.5        & \textbf{0.01}                  \\
				\textbf{LNUCB-TA}          & 1                                & \textbf{752}                & \textbf{0.94}        & 293.83       & \textbf{0.01}                 \\
				\textbf{LNUCB-TA}          & 10                               & \textbf{752}                & \textbf{0.94}        & 297.28       & \textbf{0.01}                  \\ \bottomrule
			\end{tabular}
		\end{center}
		\centering
		\caption{Comparative analysis of \textbf{LNUCB-TA} against conventional linear, nonlinear, and vanilla combination models. It contrasts LNUCB-TA's superior performance with those of solely linear (LinUCB), nonlinear ($k$-NN UCB), and basic linear-nonlinear combinations ((Lin+$k$-NN)-UCB) across various exploration rates.}
		\label{table:Comparative-analysis-of-LNUCB-TA}
	\end{table}
	
	Figure \ref{fig:my_label} provides a comparative analysis over 800 steps, showcasing cumulative and mean rewards of LNUCB-TA against different MAB models, including enhanced Epsilon Greedy, BetaThompson, and Lin Thompson models, with our adaptive $k$-NN method and a temporal attention mechanism. The mean reward graph in Figure \ref{fig:my_label}(b) provides further insights into the efficiency of the models at each step. The LNUCB-TA model has demonstrated rapid convergence to higher mean rewards, maintaining leading performance throughout the trials. In particular, while models such as $k$-NN KL-UCB and LinUCB show competitive performance initially, they do not sustain high rewards as consistently as LNUCB-TA. Additionally, the enhancements introduced through Algorithm \ref{alg:k-NN} and the attention mechanism to traditional models have resulted in performance improvements (please refer to Appendix \ref{sec:more results}). 
	
	Complementing Figure \ref{fig:my_label}, Table~\ref{table:Comparative-analysis-of-LNUCB-TA} contrasts the performance of LNUCB-TA with purely linear models (LinUCB), purely nonlinear models ($k$-NN UCB), and a basic linear-nonlinear combination ((Lin+$k$-NN)-UCB) across various exploration rates. This table demonstrates that at lower exploration rates (0.1 and 1), linear models outperform nonlinear models, whereas at a higher exploration rate (10), nonlinear models excel. The basic combination generally surpasses both linear and nonlinear models at exploration rates of 0.1 and 1 but performs worse than nonlinear models at an exploration rate of 10. However, our hybrid model, LNUCB-TA, consistently outperforms all these models at every exploration rate. It also requires less time compared to the vanilla combinations, highlighting the refined efficacy and efficiency of LNUCB-TA in dynamically adjusting to complex environments.

	\subsection{Ablation Study} 
\begin{figure}[H]
		\begin{center}
			\fbox{\includegraphics[width=0.98\textwidth]{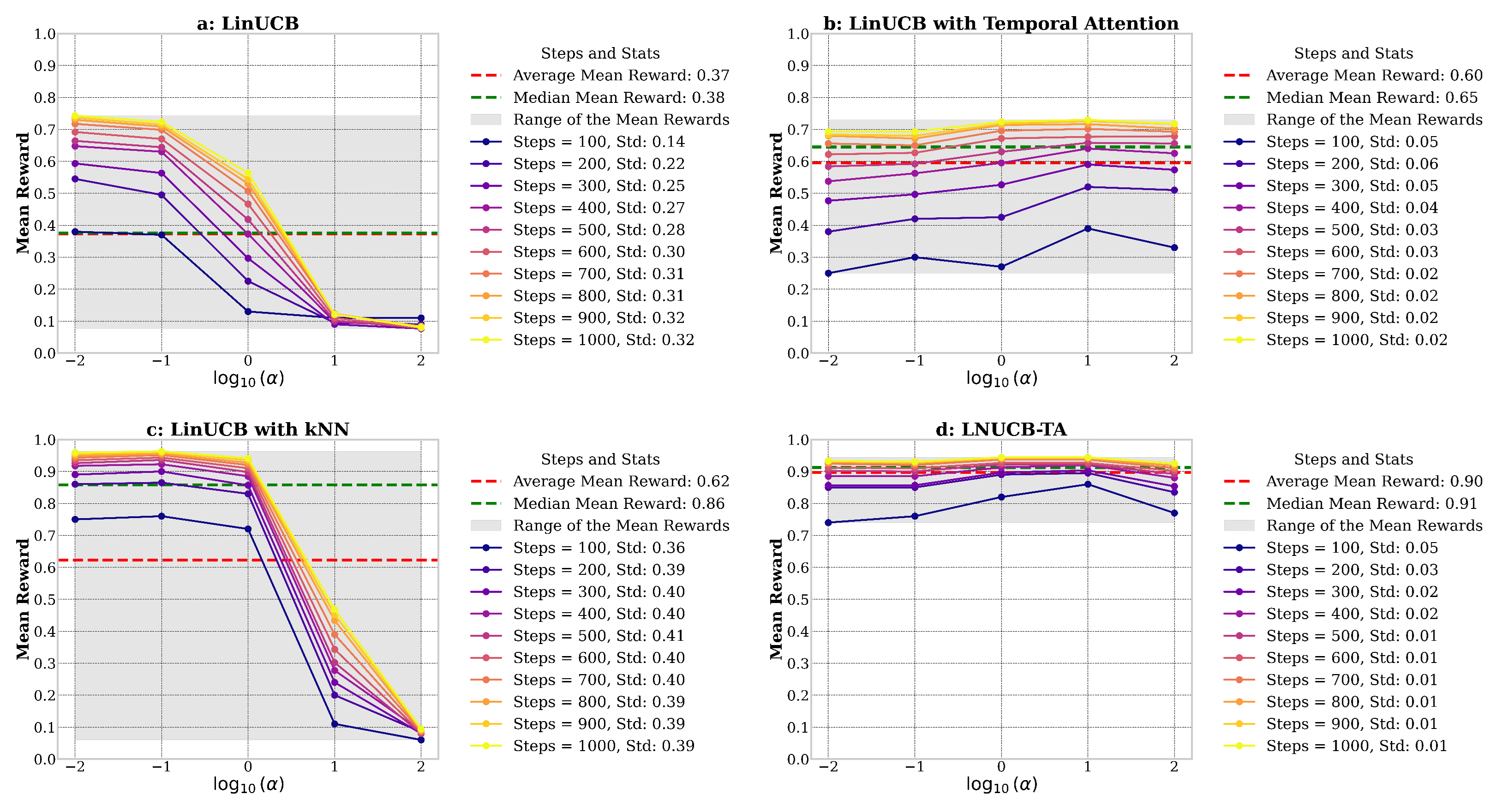}}
		\end{center}
		\caption{Ablation study: Model (a) is the base LinUCB model. Model (b) incorporates the temporal attention mechanism, significantly enhancing consistency. Model (c) implements the adaptive $k$-NN approach, increasing average mean rewards. Model (d) \textbf{LNUCB-TA} integrates both temporal attention and adaptive $k$-NN, achieving the highest average and median rewards with the greatest consistency.}
		\label{fig:ablation}
	\end{figure}

    We have assessed the impact of integrating our novel components through various model variants, as shown in Figure \ref{fig:ablation}. Model (a) represents the base LinUCB model. Model (b), which incorporates the temporal attention mechanism, significantly enhances reward consistency, reducing the standard deviation from 0.32 (Model (a)) to 0.02. This indicates that dynamic adjustment of the exploration parameter, informed by historical data relevance, effectively stabilizes reward outcomes. Model (c), which implements the adaptive $k$-NN approach, increases average mean rewards from 0.37 to 0.62 by optimizing the number of neighbors based on observed reward variance, capturing more nuanced patterns and improving prediction accuracy. While Model (b) ensures robustness against environmental fluctuations, Model (c), despite its higher average reward, exhibits greater variability. Model (d) (LNUCB-TA), integrating both temporal attention and adaptive $k$-NN, achieves the highest average mean reward (0.90) and median reward (0.91), with the greatest consistency among all models tested. This demonstrates that combining these enhancements effectively balances exploration and exploitation, setting a new standard for adaptability and precision in dynamic MAB environments.
	
	\section{Conclusion}
	\label{sec:conclusion}
	This study introduced LNUCB-TA, a hybrid contextual multi-armed bandit model that effectively integrates linear and nonlinear estimation techniques within the joint space of context features and reward history. Our findings demonstrate that the proposed adaptive $k$-NN component significantly enhances reward predictions by dynamically adjusting to evolving reward distributions and feature representations. Furthermore, the temporal attention mechanism refines the exploration-exploitation trade-off, allowing real-time adaptation of exploration factors based on data variability. These improvements contribute to a more adaptive and robust decision-making framework, capable of outperforming conventional MAB models across diverse settings. The results highlight the potential of incorporating nonlinear learning and context-aware exploration strategies in bandit problems, paving the way for future research on more flexible and data-efficient reinforcement learning methods. We also prove that the regret of LNUCB-TA is optimal up to \( R_T = O(\sqrt{ dT \log(T)}) \), demonstrating a sub-linear regret.
	
	\newpage
	\vskip 0.2in
	\bibliography{LNUCB-TA_JMLR}
	
	\newpage  
	\appendix
	\section{Appendix}
	\subsection{Proofs}
	\label{sec:Proof}

	\begin{longtable}{@{}>{\raggedright\arraybackslash}p{3cm} 
			>{\raggedright\arraybackslash}p{10cm}@{}}
		\toprule
		\textbf{Notation}        & \textbf{Representation} \\ \midrule
		\endfirsthead
		\multicolumn{2}{@{}l}{\textit{Continued from previous page}} \\ \toprule
		\textbf{Notation}        & \textbf{Representation} \\ \midrule
		\endhead
		\multicolumn{2}{@{}l}{\textit{Continued on next page}} \\ \bottomrule
		\endfoot
		\bottomrule
		\caption{Comprehensive list of notations and their representations used in the proposed approach.}
		\label{tab:notation}
		\endlastfoot
		
		$X$                      & Context feature space \\
		$Z$                      & Reward history space \\
		$\rho$                   & Metric distance function \\
		$T$                      & Total number of time steps \\
		$t$                      & Time index \\
		$x_t$                    & Context feature vector at time $t$ \\
		$z_t^a$                  & Historical rewards for arm $a$ up to time $t$ \\
		$a$                      & Index of an arm \\
		$A$                      & Set of all arms and the total number of arms $\{1, \dots, A\}$ \\
		$Y_t^a$                  & Reward random variable for arm $a$ at time $t$ \\
		$\hat{Y}_t^a$            & Observed reward for arm $a$ at time $t$ \\
		$\xi_t^a$                & Noise term capturing stochastic errors \\
		$\mu_t^a$                & Parameter vector for the linear model of arm $a$ at time $t$ \\
		$f_{k,t}^a(x_t^a, z_t^a)$ & Nonlinear component estimated by $k$-NN for arm $a$ at time $t$ \\
		$N_k^a(x_t^a)$           & Set of $k$-nearest neighbors for context $x_t^a$ based on Euclidean distance \\
		$\text{BALL}_t^a(x_t^a, r)$ & Open metric ball centered at $x_t^a$ with radius $r$ for arm $a$ \\
		$\pi$                    & Decision policy guiding arm selection \\
		$\pi_t^*$                & Optimal policy at time $t$ \\
		$H_{t-1}$                & Historical data up to time $t-1$, including contexts, chosen arms, and rewards \\
		$\alpha$                 & Exploration parameter \\
		$\alpha_0$               & Initial exploration parameter value \\
		$\alpha_{N_t^a}$         & Dynamic exploration parameter for arm $a$ based on selection frequency and reward patterns \\
		$g$                      & Global performance metric (average reward across all arms) \\
		$n_t^a$                  & Average reward history of arm $a$ up to time $t$ \\
		$N_t^a$                  & Number of pulls of arm $a$ up to time $t$ \\
		$\kappa$                 & Weighting factor balancing global ($g$) and local ($n_t^a$) attention \\
		$R_T(\pi)$               & Total regret of the policy $\pi$ over $T$ time steps \\
		$\Delta_t^a$             & Indicator function for arm $a$ being selected at time $t$ \\
		$\text{regret}_t^a$      & Instantaneous regret for arm $a$ at time $t$ \\
		$l_t^a(x_t^a)$           & Linear prediction for arm $a$ at context $x_t^a$, defined as $\mu_t^a)^\top x_t^a$ \\
		$D$                      & Decision space for context selection \\
		$g_t^a(x_t^a, z_t^a)$    & Optimal expected reward for arm $a$ based on context $x_t^a$ and rewards $z_t^a$ \\
		$UCB_t^a$                & Upper Confidence Bound for arm $a$ at time $t$, incorporating dynamic exploration \\
		$\theta_{\min}, \theta_{\max}$ & Minimum and maximum thresholds for dynamic $k$-selection in $k$-NN \\
		$\text{Var}(z_t^a)$      & Variance of reward history for arm $a$ at time $t$ \\
		$\beta_t^a$              & Confidence parameter for arm $a$ at time $t$, influencing the regret bound \\
		$d$                      & Dimensionality of the feature space \\
		$B$                      & Upper bound for the norm of context vectors \\
		$W$                      & Upper bound on the norm of the true parameter vector \\
		$\lambda$                & Regularization parameter, defined as $\sigma^2 W^2$ \\
		$\sigma$                 & Upper bound on noise in reward observation \\
		$\delta$                 & Probability threshold for maintaining confidence bounds \\
		$u_{t,k}^a$              & Uncertainty contributed by $k$-NN predictions for arm $a$ at time $t$ \\
		$e_{t,k}^a$              & Variance term contributed by $k$-NN predictions, $e_{t,k}^a = (u_{t,k}^a)^2$ \\
		$\gamma$                 & Scaling constant for the variance term $e_{t,k}^a$ \\
		$\Sigma_t^a$             & Covariance matrix updated to time $t$ for arm $a$ \\
		$w_t^a$                  & Normalized width at time $t$ for arm $a$, defined as $(x_t^a)^T (\Sigma_t^a)^{-1} x_t^a$ \\
		$\text{det}(\Sigma_T^a)$ & Determinant of the covariance matrix $\Sigma_T^a$ for arm $a$ at time $T$ \\
		$\text{Trace}(\cdot)$    & Sum of the eigenvalues or diagonal elements of a matrix \\
		$R_T$                    & Cumulative regret up to time $T$ \\
		$\log(\text{det}(\cdot))$ & Logarithm of the determinant of a matrix \\
		$\tilde{O}(\cdot)$       & Big-O notation absorbing logarithmic factors \\
		$\tau_a$                 & Reach of the manifold $M$ for arm $a$, representing stable geometric properties \\
		$c_0$                    & Constant governing regularity of the uncertainty region \\
		$\nu_a$                  & Regular measure associated with arm $a$, continuous with respect to $v_a$ \\
		$v_a$                    & Volume measure for the uncertainty region of arm $a$ \\
		$R_X^a$                  & Radius for the local neighborhood around $(x_a, z_a)$ \\
		$\|x\|$                  & Norm of the feature vector $x$ \\
		$\frac{d\alpha_{N_t^a}}{dN_t^a}$ & Rate of change of the exploration parameter $\alpha_{N_t^a}$ with respect to $N_t^a$ \\
		$\Gamma_{t,k}^a(x_t^a)$  & Set of indices corresponding to the $k$-nearest neighbors for context $x_t^a$ \\
		$\overline{Y}^a$         & Average reward for arm $a$ over all observed rewards \\
	\end{longtable}

	\paragraph{Proof sketch.}This section provides a structured and detailed exposition of the proofs for Theorems \ref{thm:regret} and \ref{thm:exploration}. For Theorem \ref{thm:regret}, the proof is comprehensive and requires the establishment of the two critical propositions \ref{prop:ucb} and \ref{prop:sum-regret}. 
	We begin with an overview of the model’s parameters and introduce definitions crucial for understanding the proofs. Next, we list the key assumptions that underpin the theorem and its supporting propositions. Following this, we detail and prove the supporting lemmas that provide the necessary groundwork for the propositions. Using these lemmas, we rigorously prove each proposition, which directly supports the final proof of Theorem \ref{thm:regret}. Finally, after proving the sub-linear regret bound (Theorem \ref{thm:regret}), we prove Theorem \ref{thm:exploration} by relying on fundamental principles of the GALA concept.
	
	\paragraph{Model overview.} As discussed in Section \ref{sec:methodology} of the paper, we have a hybrid contextual MAB problem, where the expected reward for each arm $a$ at context $x_t$ is modeled through a linear and a nonlinear component defined as equation (\ref{eq:reward}). This formulation seeks to effectively combine linear insights with the local history learned from the {$k$}-NN approach, adjusting for historical reward data $z_t$, which comprises past rewards related to arm $a$ according to Algorithm \ref{alg:k-NN}.
	Additionally, the regret associated with each arm $a$ at time $t$ quantifies the difference between the reward that could have been achieved by selecting the optimal action and the reward actually received by equation (\ref{eq:regret}). As a result, total regret is calculated as equation (\ref{eq:regret-total}). This measure of regret reflects the performance difference and highlights the effectiveness of the decision policy in approximating the optimal action choices over time.
	
	\begin{corollary}[Uncertainty region]\label{cor:uncertainty-region}
		The essence of LNUCB-TA revolves around the concept of "optimism in the face of uncertainty" \citep{liu2024ovd,kamiura2017optimism,lykouris2021corruption, li2010contextual,russo2013eluder}. Following \citep[Section 8.3]{chu2011contextual}, the center of an uncertainty region, $\text{BALL}_t^a$ is $\hat{\mu}_t^a$, which is the solution of the following ridge regression problem:
		\begin{equation}
			\begin{split}
				\hat{\mu}_t^a &= \arg\min_{\theta} \left\| (X_t^a)^T \theta - (Y_t^a - f_{k,t}^a(x_t^a, z_t^a)) \right\|_2^2 + \lambda \|\theta\|_2^2 \\
				&= ((X_t^a)^T X_t^a + \lambda I)^{-1} (X_t^a)^T (Y_t^a - f_{k,t}^a(x_t^a, z_t^a)) \\
				&= (\Sigma_t^a)^{-1} \sum_{t=0}^{t-1} X_t^a (Y_t^a - f_{k,t}^a(x_t^a, z_t^a)),
			\end{split}
		\end{equation}
		
		where \(\theta\) is the parameter vector being optimized, \(\lambda\) is the regularization parameter, and \(\Sigma_t^a = (X_t^a)^T X_t^a + \lambda I\) is the covariance matrix \citep[equation 20.1]{lattimore2020bandit} updated to time \(t\) for arm \(a\), reflecting the context feature information and the regularization term.
	\end{corollary}
	\begin{definition}\label{def:ball-definition}
		For LNUCB-TA, the shape of the region \(\text{BALL}_t^a\) following corollary \ref{cor:uncertainty-region} is defined through the feature covariance \(\Sigma_t^a\). Precisely, the uncertainty region, or \textit{confidence ball}, is defined as:
		\begin{equation}
			\text{BALL}_t^a = \{\mu \mid (\mu - \hat{\mu}_t^a)^T \Sigma_t^a (\mu - \hat{\mu}_t^a) \leq \beta_t^a\}.
		\end{equation}
	\end{definition}
	
	\begin{corollary}[Uncertainty of nonlinear estimation]\label{cor:uncertainty}
		Following \citep[Section 3.1]{reeve2018k}, for each context \(x_t^a\) in \(\mathcal{X}\) and each arm \(a\), at a given time step \(t \in [n]\) and with access to the reward history up to \(t\), represented as \(\mathcal{Z}_{t}\), we define an enumeration of indices from \([t-1]\) as \(\{\tau_{t,q}^a(x_t^a)\}_{q \in [t-1]}\) for each arm \(a\) as
		\begin{equation}
			\rho((x_t^a, z_t^a), (X_{\tau_{t,q}^a(x_t^a)}, Z_{\tau_{t,q}^a(x_t^a)})) \leq \rho((x_t^a, z_t^a), (X_{\tau_{t,q+1}^a(x_t^a)}, Z_{\tau_{t,q+1}^a(x_t^a)})).
		\end{equation}
		This enumeration is ordered such that for each \(q \leq t-2\), where \(q\) is a numeric \( \mathbb{N} \), \(X_{\tau_{t,q}^a(x_t^a)}\) and \(Z_{\tau_{t,q}^a(x_t^a)}\) are the historical contexts and rewards associated with arm \(a\) at index \(\tau_{t,q}^a(x_t^a)\).
		Given \(k \in [t-1]\), \(\Gamma_{t,k}^a(x_t^a)\) is defined as
		\begin{equation}
			\Gamma_{t,k}^a(x_t^a) = \{\tau_{t,q}^a(x_t^a) : q \in [k]\} \subseteq [t-1].
		\end{equation}
		This set includes indices of the \(k\) closest historical data points to the current feature vector \(x_t^a\) for arm \(a\), selected based on their proximity in the combined feature and reward space as measured by \(\rho\).
		The maximum distance or uncertainty measure for arm \(a\) at time \(t\), \(u_{t,k}^a(x_t^a)\), satisfies
		\begin{equation}
			u_{t,k}^a = \max\{\rho((x_t^a, z_t^a), (X_s, Z_s^a)) : s \in \Gamma_{t,k}^a(x_t^a)\} = \rho((x_t^a, z_t^a), (X_{\tau_{t,k}^a(x_t^a)}, Z_{\tau_{t,k}^a(x_t^a)})).
		\end{equation}
		This measure assesses the greatest distance between the current feature vector and reward data \((x_t^a, z_t^a)\) and those of the historical data within the nearest neighbors.
	\end{corollary}

	\begin{corollary}[Arm-specific regular sets and measures]\label{cor:regular-set}
		Using \citep[Definition 1]{reeve2018k}, we can state that in the extended metric space \((\mathcal{X} \times \mathcal{Z}, \rho)\), where \(\mathcal{X} \times \mathcal{Z}\) represents the joint space of context features and reward history for arm \(a\), and \(\rho\) is the metric, a subset \(A \subset \mathcal{X} \times \mathcal{Z}\) is a \((c_0, r_0^a)\) regular set if for all \((x^a, z^a) \in A\) and all \(r \in (0, r_0^a)\),
		\begin{equation}
			v^a (A \cap \text{BALL}^a((x^a, z^a); r)) \geq c_0 \cdot v^a (\text{BALL}^a((x^a, z^a); r)).
		\end{equation}
		A measure \(\nu^a\) with \(\operatorname{supp}(\nu^a) \subset \mathcal{X} \times \mathcal{Z}\) is a \((c_0, r_0^a, \nu_{\text{min}}^a, \nu_{\text{max}}^a)\) regular measure with respect to \(v^a\) if \(\operatorname{supp}(\nu^a)\) is a \((c_0, r_0^a)\)-regular set with respect to \(v^a\) and \(\nu^a\) is absolutely continuous with respect to \(v^a\) with Radon-Nikodym derivative \citep[Theorem 3.8]{folland1999real} as
		\begin{equation}
			v^a (x^a, z^a) = \frac{d\nu^a (x^a, z^a)}{dv^a (x^a, z^a)},
		\end{equation}
		ensuring
		\begin{equation}
			\nu_{\text{min}}^a \leq v^a (x^a, z^a) \leq \nu_{\text{max}}^a.
		\end{equation}
	\end{corollary}
	\begin{assumption}[Arm-specific dimension assumption]\label{assumption:dimension}
		Applying \citep[Section 2.2]{rigollet2010nonparametric}, we can assume that for each arm \(a \in \{1, \ldots, A\}\), there exist constants \(C_d\), \(d\), and \(R_X^a > 0\) such that for all \((x^a, z^a) \in \operatorname{supp}(\nu^a)\) and \(r \in (0, R_X^a)\), it holds
		\begin{equation}
			\nu^a(\text{BALL}^a((x^a, z^a); r)) \geq C_d \cdot r^d.
		\end{equation}
		Here, \(r\) represents the radius of the ball in the joint space of features and reward history for arm \(a\), indicating the scale of the local neighborhood around \((x^a, z^a)\) considered for the measure. The \(\text{BALL}^a((x^a, z^a); r)\) highlights the dependency on both context and past rewards within this radius.
		To prove this assumption, we shall follow a corollary followed from \citep[Lemma 12]{eftekhari2015new}.
	\end{assumption}
	\begin{corollary} [Arm-specific dimension] \label{cor:corr-1}
		For each arm \(a\), let \(M \subseteq \mathbb{R}^{D}\) be a \(C^\infty\)-smooth compact sub-manifold of uniform dimension \(d\) \citep{lee2006riemannian} with a defined reach \(\tau^a\) \citep{federer1959curvature}, quantified based on \citep{niyogi2008finding} as
		\begin{equation}
			\tau^a := \sup \left\{ r > 0 : \forall j \in \mathbb{R}^{D}, \inf_{q \in M} \left\{ \|j - q\|_2 \right\} < r \implies \exists! \, p \in M, \|j - p\|_2 = \inf_{q \in M} \left\{ \|j - q\|_2 \right\} \right\}.
		\end{equation}
		This reach reflects the maximum radius such that for every point $j$ within this distance from the manifold $M$, there is a nearest point on the manifold, ensuring stable local geometric properties, supported by \citep[Lemma 7.2]{boissonnat2018geometric}. If \(\nu^a\) is a \((c_0, R_0^a, \nu_{\text{min}}^a, \nu_{\text{max}}^a)\) regular measure with respect to \(V_{M}\), then \(\nu^a\) satisfies assumption \ref{assumption:dimension} with constants \(R_X^a = \min\{\tau^a / 4, R_0^a\}\), \(d\), and \(C_d = \nu_{\text{min}}^a \cdot c_0 \cdot v_d^a \cdot 2^{-d}\), where \(v_d^a\) is the Lebesgue measure of the unit ball in \(\mathbb{R}^d\).
	\end{corollary}
	
	\textit {Proof.} For each arm $a$, consider any point $(x^a, z^a) \in \operatorname{supp}(\nu^a)$ and radius $r \in (0, R_X^a)$. Applying the \citep[Lemma 5.3]{niyogi2008finding}, for arm $a$, the volume within the $\operatorname{BALL}_r((x^a, z^a))$ can be estimated by
	\begin{equation}
		V_M(\operatorname{BALL}_r((x^a, z^a))) \geq \left(1 - \frac{r^2}{4(\tau^a)^2}\right)^{\frac{d}{2}} \cdot v_d \cdot r^d.
	\end{equation}
	This equation reflects the geometrical properties of the manifold within a local neighborhood around $(x^a, z^a)$, given the manifold's reach and dimensionality.
	
	Moreover, since $\nu^a$ is $(c_0, R_0^a, \nu_{\text{min}}^a, \nu_{\text{max}}^a)$-regular using corollary \ref{cor:regular-set}, it holds
	\begin{equation}
		\nu^a(\operatorname{BALL}_r((x^a, z^a))) \geq \nu_{\text{min}}^a \cdot c_0 \cdot V_M(\operatorname{BALL}_r((x^a, z^a)))
	\end{equation}
	Combining this with the volume estimation provided by corollary \ref{cor:corr-1}, we get
	\begin{equation}
		\nu^a(\operatorname{BALL}_r((x^a, z^a))) \geq \nu_{\text{min}}^a \cdot c_0 \cdot \left(1 - \frac{r^2}{4(\tau^a)^2}\right)^{\frac{d}{2}} \cdot v_d \cdot r^d,
	\end{equation}
	\begin{equation}
		\nu^a(\operatorname{BALL}_r((x^a, z^a))) \geq \nu_{\text{min}}^a \cdot c_0 \cdot v_d \cdot 2^{-d} \cdot r^d, 
		\label{eq:eq-29}
	\end{equation}.
	
	This calculation demonstrates that the measure $\nu^a$ within the ball $\operatorname{BALL}_r((x^a, z^a))$ exceeds a lower bound that scales with $r^d$, the dimensionally-scaled radius of influence. This establishes the local density and regularity of $\nu^a$ around each point in its support, confirming the validity of the arm-specific dimension assumption for the manifold $M$.
	
	\begin{assumption}[Bounded rewards assumption]\label{assumption:bounded-reward}
		For all time steps $t \in [n]$ and for each arm $a \in [A]$, the rewards $Y_t^a$ observed after integrating both linear and {$k$}-Nearest Neighbors ({$k$}-NN) adjustments are bounded within an interval assumed as
		\begin{equation}
			-1 \leq Y_t^a \leq 1.
		\end{equation}
	\end{assumption}
	\begin{assumption}[Confidence in parameter estimation]\label{assumption:miu-in-ball}
		For all time steps $t \in [n]$ and for each arm $a \in [A]$, we shall assume that the true parameter vector $\mu^*$ resides within a confidence ball centered around the estimated parameter $\mu_t^a$. This confidence ball, denoted as $\text{BALL}_{(t,a)}$, is defined based on the estimation error and the uncertainty in the measurements up to time $t$, incorporating adjustments for both linear and nonlinear adjustments.
	\end{assumption}
	\begin{lemma}[Width of confidence Ball for LNUCB-TA]\label{lem:Width-ball}
		Let \( x \in D \). As \( \mu \) belongs to \( \text{BALL}_t^a \) for each arm \( a \) and \( x \in D \) according to assumption \ref{assumption:miu-in-ball}, then
		\begin{equation}
			|\left( \mu - \hat{\mu}_t^a \right)^T x| \leq \sqrt{\beta_t^a x^T (\Sigma_t^a)^{-1} x}.
		\end{equation}
		This lemma follows \citep[Lemma 6.8]{agarwal2019reinforcement}.
	\end{lemma}
	
	\textit {Proof.} Starting with the absolute value of the dot product of \((\mu - \hat{\mu}_t^a)\) and \(x\), we get
	\begin{equation}
		|\left(\mu - \hat{\mu}_t^a\right)^T x|.
	\end{equation}
	By utilizing the Cauchy-Schwarz inequality \citep[Section 1.2]{strang2022introduction}, which states that for all vectors \(u\) and \(v\) in an inner product space, we have
	\begin{equation}
		|\langle u, v \rangle|^2 \leq \langle u, u \rangle \cdot \langle v, v \rangle,
	\end{equation}
	where \(\langle \cdot, \cdot \rangle\) is the inner product. Every inner product gives rise to a Euclidean \(l_2\) norm, called the canonical or induced norm, where the norm of a vector \(u\) is defined by
	\begin{equation}
		\|u\| := \sqrt{\langle u, u \rangle}.
		\label{eq:uv}
	\end{equation}
	By taking the square root of both sides of equation(\ref{eq:uv}), the Cauchy-Schwarz inequality can be written in terms of the norm
	\begin{equation}
		|\langle u, v \rangle| \leq \|u\| \|v\|.
	\end{equation}
	Moreover, the two sides are equal if and only if $u$ and $v$ are linearly dependent. Applying this inequality to $u = (\Sigma_t^a)^{1/2} (\mu - \hat{\mu}_t^a)$ and $v = (\Sigma_t^a)^{-1/2} x_t^a$, we get
	\begin{equation}
		\begin{split}
			|(\mu - \hat{\mu}_t^a)^T x| &= |((\Sigma_t^a)^{1/2} (\mu - \hat{\mu}_t^a))^T (\Sigma_t^a)^{-1/2} x| 
			\leq \|(\Sigma_t^a)^{1/2} (\mu - \hat{\mu}_{t,a})\| \cdot \|(\Sigma_t^a)^{-1/2} x\| \\
			&= \sqrt{(\mu - \hat{\mu}_t^a)^T \Sigma_t^a (\mu - \hat{\mu}_t^a)} \cdot \sqrt{x^T (\Sigma_t^a)^{-1} x}.
		\end{split}\label{eq:back}
	\end{equation}
	Since \( \mu \) is assumed to be within the confidence set \( \text{BALL}_t^a \) as assumption \ref{assumption:miu-in-ball}, we have
	\begin{equation}
		(\mu - \hat{\mu}_t^a)^T \Sigma_t^a (\mu - \hat{\mu}_t^a) \leq \beta_t^a,
	\end{equation}
	plugging this back into equation (\ref{eq:back}), we can obtain
	\begin{equation}
		|(\mu - \hat{\mu}_t^a)^T x| \leq \sqrt{\beta_t^a} \cdot \sqrt{x^T (\Sigma_t^a)^{-1} x} = \sqrt{\beta_t^a x^T (\Sigma_t^a)^{-1} x}.
	\end{equation}
	
\begin{lemma}[Normalized width for LNUCB-TA)]\label{lem:normalized-width}
	Fix $t \leq T$. As $(\mu^a)^* \in \text{BALL}_t^a$ based on assumption \ref{assumption:miu-in-ball}, we define
	\begin{equation}
		w_t^a = \sqrt{(x_t^a)^T (\Sigma_t^a)^{-1} x_t^a},
	\end{equation}
	which is the "normalized width" at time $t$ for arm $a$ in the direction of the chosen decision, then
	\begin{equation}
		\text{regret}_t^a \leq 2 \min\left(\sqrt{\beta_t^a} w_t^a, 1\right) \leq 2 \sqrt{\beta_T^a} \min(w_t^a, 1).
	\end{equation}
\end{lemma}
This lemma is inspired by the theoretical analysis of nonlinear bandits presented in \citep{dong2021provable}, where the sample complexity for finding an approximate local maximum is discussed, leveraging the model complexity rather than the action dimension. Additionally, the approach to handling confidence bounds in linear bandits \citep{agrawal2013thompson,li2010contextual}, provides a foundational understanding for the linear components of this work.

\textit {Proof.} Let $\tilde{\mu} \in \text{BALL}_t^a$, wedefine instantaneous regret as
\begin{equation}
	\begin{split}
		\text{regret}_t^a &= (\mu^a)^T {(x^a)}^* - (\mu^a)^T x_t^a \leq (\tilde{\mu} - (\mu^a)^*)^\top x_t^a \\
		&= (\tilde{\mu} - \hat{\mu}_t^a)^\top x_t^a + (\hat{\mu}_t^a - (\mu^a)^*)^\top x_t^a.
	\end{split}
\end{equation}
For the sum of two inner products, the triangle inequality \citep[Section 4.5]{axler2015linear} gives
\begin{equation}
	|(\tilde{\mu} - \hat{\mu}_t^a)^\top x_t^a + (\hat{\mu}_t^a - (\mu^a)^*)^\top x_t^a| \leq |(\tilde{\mu} - \hat{\mu}_t^a)^\top x_t^a| + |(\hat{\mu}_t^a - (\mu^a)^*)^\top x_t^a|,
\end{equation}

and by using the given bound for $|(\mu - \hat{\mu}_t^a)^\top x|$ in lemma \ref{lem:Width-ball}, we can obtain
\begin{equation}
	|(\tilde{\mu} - \hat{\mu}_t^a)^\top x_t^a| \leq \sqrt{\beta_t^a (x_t^a)^T (\Sigma_t^a)^{-1} x_t^a} = \sqrt{\beta_t^a} w_t^a,
\end{equation}
\begin{equation}
	|(\hat{\mu}_t^a - (\mu^a)^*)^\top x_t^a| \leq \sqrt{\beta_t^a (x_t^a)^T (\Sigma_t^a)^{-1} x_t^a} = \sqrt{\beta_t^a} w_t^a.
\end{equation}

Thus, 
\begin{equation}
	|(\tilde{\mu} - \hat{\mu}_t^a)^\top x_t^a + (\hat{\mu}_t^a - (\mu^a)^*)^\top x_t^a| \leq 2 \sqrt{\beta_t^a} w_t^a,
\end{equation}

and since $-1 \leq Y_t^a \leq 1$ (assumption \ref{assumption:bounded-reward}), the regret is at most 2, then
\begin{equation}
	\text{regret}_t^a \leq 2 \sqrt{\beta_t^a} w_t^a \leq \min(2 \sqrt{\beta_t^a} w_t^a, 2).
\end{equation}

Expressing it with 2 outside the minimum function for clarity and to align with the bound mentioned in assumption \ref{assumption:bounded-reward}, satisfies
\begin{equation}
	\text{regret}_t^a \leq 2 \min(\sqrt{\beta_t^a} w_t^a, 1),
\end{equation}

and as $\beta_t^a$ is non-decreasing over time (common in learning systems where confidence typically increases with more data), $\beta_T^a \geq \beta_t^a$ for any $t \leq T$. Thus, applying this monotonicity property of $\beta_t^a$,
\begin{equation}
	2 \sqrt{\beta_t^a} \min(w_t^a, 1) \leq 2 \sqrt{\beta_T^a} \min(w_t^a, 1),
\end{equation}

which completes the proof.

\begin{lemma}[Determinant expansion]\label{lem:deteminanat}
	We have
	\begin{equation}
		\text{det}(\Sigma_{T}^a) = \text{det}(\Sigma_{0}^a) \prod_{t=0}^{T-1} (1 + (w_t^a)^2 + \gamma e_{t,k}^a ),
	\end{equation}
	where \( e_{t,k}^a  = \left(u_{t,k}^a{}\right)^2 \). This lemma is structured based on \citep[Lemma 6.8]{agarwal2019reinforcement} and \citep[Theorem 1]{perrault2020covariance}.
\end{lemma}

\textit {Proof.} By definition of $\Sigma_{t+1}^a$, we get
\begin{equation}
	\Sigma_{t+1}^a = \Sigma_t^a + x_t^a (x_t^a)^\top + \gamma e_{t,k}^a  I,
\end{equation}
where $\gamma$ helps to scale the identity matrix $I$ multiplied by the variance term $e_{t,k}^a $, which quantifies the uncertainty contributed by the {$k$}-NN predictions at each time step for arm $a$.
Considering the determinant, we have
\begin{equation}
	\text{det}(\Sigma_{t+1}^a) = \text{det}(\Sigma_t^a + x_t^a (x_t^a)^\top + \gamma e_{t,k}^a  I).\label{eq:sigma}
\end{equation}

Then, a special case of "matrix determinant lemma" attributed to \citep[Corollary 18.2.10]{harville1998matrix}, originated from \citep{sherman1950adjustment} is applied as equation (\ref{eq:matrix})
\begin{equation}
	\text{det}(A + uv^\top) = \text{det}(A)(1 + v^\top A^{-1} u)
	\label{eq:matrix}.
\end{equation}
Applying the concept of equation (\ref{eq:matrix}) in equation (\ref{eq:sigma}), we can obtain
\begin{equation}
	\text{det}(\Sigma_{t+1}^a) = \text{det}(\Sigma_t^a) \text{det}\left(I + (\Sigma_t^a)^{-1/2} x_t^a (x_t^a)^\top (\Sigma_t^a)^{-1/2} + \gamma e_{t,k}^a  (\Sigma_t^a)^{-1/2} I (\Sigma_t^a)^{-1/2}\right).
\end{equation}
Then, by decomposing the calculation further, considering $v_t = (\Sigma_t^a)^{-1/2} x_t^a$ and $u_t = \gamma e_{t,k}^a  I$,
\begin{equation}
	\text{det}(I + v_t v_t^\top + \gamma e_{t,k}^a  (\Sigma_t^a)^{-1/2} I (\Sigma_t^a)^{-1/2}) = \text{det}(I + v_t v_t^\top + \gamma e_{t,k}^a  I).
\end{equation}
Since $I$ is the identity matrix and commutes with any matrix, using the property that $\Sigma_t^a{}^{-1/2} I \Sigma_t^a{}^{-1/2} = I$ due to normalization, and where $v_t^a = (\Sigma_t^a)^{-1/2} x_t^a$ based on the proof of lemma \ref{lem:normalized-width}. Now we can observe $(v_t^a)^\top v_t^a = (w_t^a)^2$ and
\begin{equation}
	(I + v_t^a (v_t^a)^\top) v_t^a = v_t^a + v_t^a ((v_t^a)^\top v_t^a) = (1 + (w_t^a)^2) v_t^a.
\end{equation}
For this reason \( (1 + (w_t^a)^2) \) is an eigenvalue of \( I + v_t^a (v_t^a)^\top \). Since \( v_t^a (v_t^a)^\top \) is a rank one matrix, all other eigenvalues of \( I + v_t^a (v_t^a)^\top \) equal 1. Hence, \(\text{det}(I + v_t^a (v_t^a)^\top) = (1 + (w_t^a)^2)\), is implies
\begin{equation}
	\text{det}(I + v_t^a (v_t^a)^\top + \gamma e_{t,k}^a  I) = \text{det}(I + (w_t^a)^2 + \gamma e_{t,k}^a ),
\end{equation}
which gets
\begin{equation}
	\text{det}(\Sigma_{t+1}^a) = (1 + (w_t^a)^2 + \gamma e_{t,k}^a )\text{det}(\Sigma_t^a).\label{eq:iter}
\end{equation}
Finally, iterating equation (\ref{eq:iter}) from \( t = 0 \) to \( T-1 \) gives
\begin{equation}
	\text{det}(\Sigma_{T}^a) = \text{det}(\Sigma_{0}^a) \prod_{t=0}^{T-1} (1 + (w_t^a)^2 + \gamma e_{t,k}^a ).
\end{equation}

\begin{lemma}[Potential function bound]\label{lem:potential-bound}
	Consider the sequence \(x_{0}^a, \ldots, x_{T-1}^a\) such that \(\|x_t^a\|_2 \leq B\) for all \(t < T\), the potential function bound is given by
	\begin{equation}
		\begin{split}
			&\log\left(\frac{\text{det}(\Sigma_{T-1}^a)}{\text{det}(\Sigma_{0}^a)}\right) = \log\left(\text{det}\left(I + \frac{1}{\lambda} \left(\sum_{t=0}^{T-1} x_t^a (x_t^a)^\top + \sum_{a=1}^A \gamma e_{t,k}^a  I\right)\right)\right) \\
			&= \log\left(\text{det}\left(I + \frac{1}{\lambda} \left(\sum_{t=0}^{T-1} x_t^a (x_t^a)^\top + \sum_{a=1}^A \gamma (u_{t,k}^a)^2 I\right)\right)\right) \\
			&\leq d \log\left(1 + \frac{1}{d\lambda} \left(TB^2 + \sum_{a=1}^A T^a (u_{t,k}^a)^2\right)\right).
		\end{split}
	\end{equation}
\end{lemma}

\textit {Proof.} For $\Sigma_{T-1}^a$, we have
\begin{equation}
	\Sigma_{T-1}^a = \Sigma_{0}^a + \sum_{t=0}^{T-1} x_t^a (x_t^a)^\top + \sum_{a=1}^A \gamma (u_{t,k}^a)^2 I.
\end{equation}
Then, we use the identity that relates the determinant of a sum to the product of eigenvalues
\begin{equation}
	\log\left(\frac{\text{det}(\Sigma_{T-1}^a)}{\text{det}(\Sigma_{0}^a)}\right) = \log\left(\text{det}\left(I + (\Sigma_{0}^a)^{-1} \left(\sum_{t=0}^{T-1} x_t^a (x_t^a)^\top + \sum_{a=1}^A \gamma (u_{t,k}^a)^2 I\right)\right)\right),
\end{equation}
which simplifies to
\begin{equation}
	\log\left(\text{det}\left(I + \frac{1}{\lambda} \left(\sum_{t=0}^{T-1} x_t^a (x_t^a)^\top + \sum_{a=1}^A \gamma (u_{t,k}^a)^2 I\right)\right)\right).
\end{equation}
Let $\sigma_1, \ldots, \sigma_d$ be the eigenvalues of $\sum_{t=0}^{T-1} x_t^a (x_t^a)^\top + \sum_{a=1}^A \gamma (u_{t,k}^a)^2 I$. Applying the Arithmetic Mean-Geometric Mean (AM-GM) Inequality \citep[Theorem 2.1]{cvetkovski2012inequalities}, we can obtain
\begin{equation}
	\text{Trace}\left(\sum_{t=0}^{T-1} x_t^a (x_t^a)^\top + \sum_{a=1}^A \gamma (u_{t,k}^a)^2 I\right) = \sum_{t=0}^{T-1} \|x_t^a\|^2 + A \gamma (u_{t,k}^a)^2.
	\label{eq:trace}
\end{equation}
Then, we shall assume $\sum_{t=0}^{T-1} \|x_t^a\|^2 \leq TB^2$, and by summing the regularizing terms, we get
\begin{equation}
	\sum_{i=1}^d \sigma_i \leq TB^2 + \sum_{a=1}^A \gamma (u_{t,k}^a)^2.
\end{equation}
Finally, using the equation (\ref{eq:trace}),
\begin{equation}
	\log\left(\text{det}\left(I + \frac{1}{\lambda} \left(\sum_{t=0}^{T-1} x_t^a (x_t^a)^\top + \sum_{a=1}^A \gamma (u_{t,k}^a)^2 I\right)\right)\right) 
\end{equation}
\begin{equation}
	= \log\left (\prod_{i=1}^d \left(1 + \frac{\sigma_i}{\lambda}\right)\right)
\end{equation}
\begin{equation}
	= \sum_{i=1}^d \log\left(1 + \frac{\sigma_i}{\lambda}\right) \leq d \log\left(1 + \frac{1}{d\lambda} \left(TB^2 + \sum_{a=1}^A \gamma (u_{t,k}^a)^2\right)\right).
\end{equation}
This inequality uses the AM-GM inequality in the form \mbox{$\log\left(\prod_{i=1}^d \left(1 + \frac{\sigma_i}{\lambda}\right)\right) \leq d\log\left(1 + \frac{\text{Trace}}{d\lambda}\right)$}.

\begin{lemma}[Linear Operator]\label{lem:linear-operator}
	Let $\Sigma_0^a$ be an initial covariance matrix, $x_t^a$ a feature vector for arm $a$ at time $t$, and $\gamma$ a scaling constant, and $u_{t,k}^a$ is defined as stated in the description. The operator
	\begin{equation}
		\Sigma_{T-1}^a = \Sigma_0^a + \sum_{t=0}^{T-1} x_t^a (x_t^a)^\top + \gamma \sum_{a=1}^A (u_{t,k}^a)^2 I
	\end{equation}
	is a linear operator from $\mathbb{R}^d$ to $\mathbb{R}^d$, where $d$ is the dimension of the feature vectors.
\end{lemma}
\textit {Proof.} A linear operator in the context of linear algebra is a mapping $L: V \to W$ between two vector spaces $V$ and $W$ that satisfies the linearity conditions \citep{rudin1964principles}:
\begin{itemize}
	\item \textbf{Additivity:} $L(u + v) = L(u) + L(v)$ for any vectors $u, v \in V$.
	\item \textbf{Homogeneity:} $L(\alpha u) = \alpha L(u)$ for any scalar $\alpha$ and vector $u \in V$.
\end{itemize}

\textbf{Additivity:}
For any vectors $u, v \in \mathbb{R}^d$,
\begin{equation}
	\Sigma_{T-1}^a(u + v) = \Sigma_0^a(u + v) + \sum_{t=0}^{T-1} x_t^a (x_t^a)^\top(u + v) + \gamma \sum_{a=1}^A (u_{t,k}^a)^2 I(u + v)
\end{equation}
\begin{equation}
	= \Sigma_0^a(u) + \Sigma_0^a(v) + \sum_{t=0}^{T-1} x_t^a ((x_t^a)^\top u + (x_t^a)^\top v) + \gamma \sum_{a=1}^A (u_{t,k}^a)^2 (Iu + Iv)
\end{equation}
\begin{equation}
	= \Sigma_0^a(u) + \sum_{t=0}^{T-1} x_t^a (x_t^a)^\top u + \gamma \sum_{a=1}^A (u_{t,k}^a)^2 Iu + \Sigma_0^a(v) + \sum_{t=0}^{T-1} x_t^a (x_t^a)^\top v + \gamma \sum_{a=1}^A (u_{t,k}^a)^2 Iv
\end{equation}
\begin{equation}
	= \Sigma_{T-1}^a(u) + \Sigma_{T-1}^a(v).
\end{equation}

\textbf{Homogeneity:}
For any scalar $\alpha$ and vector $u \in \mathbb{R}^d$,
\begin{equation}
	\Sigma_{T-1}^a(\alpha u) = \Sigma_0^a(\alpha u) + \sum_{t=0}^{T-1} x_t^a (x_t^a)^\top(\alpha u) + \gamma \sum_{a=1}^A (u_{t,k}^a)^2 I(\alpha u)
\end{equation}
\begin{equation}
	= \alpha \Sigma_0^a(u) + \alpha \sum_{t=0}^{T-1} x_t^a (x_t^a)^\top u + \alpha \gamma \sum_{a=1}^A (u_{t,k}^a)^2 Iu
\end{equation}
\begin{equation}
	= \alpha (\Sigma_0^a(u) + \sum_{t=0}^{T-1} x_t^a (x_t^a)^\top u + \gamma \sum_{a=1}^A (u_{t,k}^a)^2 Iu)
\end{equation}
\begin{equation}
	= \alpha \Sigma_{T-1}^a(u).
\end{equation}
Since $\Sigma_{T-1}^a$ satisfies both additivity and homogeneity, it is a linear operator. Hence, the lemma \ref{lem:linear-operator} is proved.

\begin{corollary} [Self-normalized bound]\label{cor:self-normalized}
For each arm \( a \), the reward is generated as
\begin{equation}
	Y_t^a = l_t^a + f_{k,t}^a(x_t^a, z_t^a) + \xi_{t}^a = \mu_t^a)^\top x_t^a + \text{{$k$}-NN}_{k,t}^a(x_t^a, z_t^a)+ \xi_{t}^a.\label{eq:rereward}
\end{equation}
Here, \(\xi_{t}^a\) is the noise term associated with arm \( a \), which captures the inherent randomness in the rewards after accounting for both the linear model's predictions and the {$k$}-NN adjustments. This term remains conditionally \(\delta\)-sub-Gaussian.

Given the linear operator proved in lemma \ref{lem:linear-operator}, the self-normalized bound, structured by \citep[Theorem 1]{abbasi2011improved} and \citep{auer2002nonstochastic}, with the probability at least \(1-\delta\) is followed by
\begin{equation}
	\left\|\sum_{t=1}^T X_{t}^a \xi_{t}^a \right\|_{(\Sigma_{t}^a)^{-1}}^2 \leq \sigma^2 \log\left(\frac{\det(\Sigma_{t}^a) \det(\Sigma_{0}^a)^{-1}}{\delta^2}\right),
\end{equation}
where \(\xi_{t}^a\) encapsulates both inherent randomness and any deviation from {$k$}-NN estimates.
\end{corollary}

\subsubsection{Proof of Theorem \ref{thm:regret}}
\paragraph{Proof of proposition \ref{prop:ucb}.} Consider the defined reward for each arm \( a \) in equation (\ref{eq:rereward}) in corollary \ref{cor:self-normalized}, the deviation of the estimated parameter \( \mu_t^a \) from the true parameter \( (\mu^{a})^* \) is calculated as
\begin{equation}
\mu_t^a - \left(\mu^{a}\right)^* = \Sigma_t^a{}^{-1} \left(\sum_{t=0}^{t-1} x_t^a \left(\left(\mu^{a}\right)^* + \xi_t^a + \text{{$k$}-NN}_{k,t}^a(x_t^a, z_t^a)\right) x_t^a - \lambda \Sigma_t^a{}^{-1} \left(\left(\mu^{a}\right)^*\right)\right).
\end{equation}

By utilizing lemma \ref{lem:normalized-width}, we can obtain
\begin{equation}
\sqrt{(\mu_t^a - \left(\mu^{a}\right)^*)^\top \Sigma_t^a (\mu_t^a - \left(\mu^{a}\right)^*)} = \|\Sigma_t^a{}^{1/2} (\mu_t^a - \left(\mu^{a}\right)^*)\|
\end{equation}

\begin{equation}
\leq \|\lambda \Sigma_t^a{}^{-1/2} \left(\mu^{a}\right)^*\| + \|\Sigma_t^a{}^{-1/2} \sum_{t=0}^{t-1} \xi_t^a x_t^a\| 
\end{equation}

\begin{equation}
\leq \sqrt{\lambda} \|\left(\mu^{a}\right)^*\| + \sqrt{2\sigma^2 \log\left(\frac{\det(\Sigma_t^a) \det(\Sigma_0)^{-1}}{\delta}\right)}.
\end{equation}

Using the triangle inequality and considering \(\Sigma_t^a{}^{-1}\) as always positive definite, implying \((\Sigma_t^a{})^{-1} \geq \frac{1}{\lambda} I\). Our goal is to lower bound \(\Pr(\forall t; \left(\mu^{a}\right)^* \in \text{BALL}_t^a)\). At \( t = 0 \), by our initial choice, \(\text{BALL}_0^a\) contains \(\left(\mu^{a}\right)^*\), hence \(\Pr(\left(\mu^{a}\right)^* \notin \text{BALL}_0^a) = 0\). For \( t \geq 1 \), we designate the failure probability for the \( t \)-th event as

\begin{equation}
\delta_t = \left(\frac{3}{\pi^2}\right) \frac{1}{t^2} \cdot 2\delta.
\end{equation}
Using the preceding results and a union bound, gives us an upper bound on the cumulative failure probability as 
\begin{equation}
1 - \Pr(\forall t; \left(\mu^{a}\right)^* \in \text{BALL}_t^a) = \Pr(\exists t; \left(\mu^{a}\right)^* \notin \text{BALL}_t^a) \leq \sum_{t=1}^\infty \left(\frac{1}{t^2} - \frac{3}{2^t}\right) 2\delta = \frac{1}{2} \cdot 2\delta = \delta.
\end{equation}

\paragraph{Proof of Proposition \ref{prop:sum-regret}.}

Considering assumption \ref{assumption:miu-in-ball} for all time steps \( t \) and arms \( a \), we start by expressing the sum of squared regrets

\begin{equation}
\sum_{t=0}^{T-1} (\text{regret}_t^a)^2  \leq \sum_{t=0}^{T-1} 4\beta_{t}^a \min((w_t^a)^2,1) \label{eq:a1}
\end{equation}
\begin{equation}
\leq 4\beta_{T}^a \sum_{t=0}^{T-1} \min((w_t^a)^2,1) 
\leq \max\{8, \frac{4}{\log 2}\} \beta_T^a\sum_{t=0}^{T-1} \log(1+(w_t^a)^2+\gamma (u_{t,k}^a)^2)
\label{eq:a2}
\end{equation}
\begin{equation}
\leq 8\beta_{T}^a \log\left(\frac{\det(\Sigma_{T-1}^a)}{\det(\Sigma_{0}^a)}\right) = 8\beta_{T}^a d\log\left(1+\frac{TB^2}{d\lambda}+\frac{\sum_{a=1}^A \sum_{t=0}^{T-1} (u_{t,k}^a)^2}{d\lambda}\right)
\label{eq:a3}
\end{equation}

The first inequality follows from lemma \ref{lem:normalized-width}. The second is from since \(\beta_{t}^a\) is an increasing function of \( t \), \(\beta_{t}^a \leq \beta_{t+1}^a\) for all \( t \) where \( 0 \leq t < T-1 \) and \(\sum_{t=0}^{T-1} \beta_{t}^a = \beta_{T}^a\).
The third follows that for \( 0 \leq y \leq 1 \), the inequality \( y \geq \log(1+y) \geq \frac{y}{1+y} \geq \frac{y}{2} \) holds, and specifically for \( (w_t^a)^2 \) within these bounds, we have
\begin{equation}
(w_t^a)^2+\gamma (u_{t,k}^a{})^2 \leq 2\log(1+(w_t^a)^2+\gamma (u_{t,k}^a{})^2 ).\label{eq:eq6}
\end{equation}
When \( (w_t^a)^2 > 1 \), the relationship shifts to
\begin{equation}
4\beta_{T}^a = \frac{4}{\log 2} \beta_{T}^a \log 2 \leq \frac{4}{\log 2} \beta_{T}^a \log(1+(w_t^a)^2+\gamma (u_{t,k}^a{})^2 ).\label{eq:eq7}
\end{equation}
The equation (\ref{eq:eq6}) follow lemma \ref{lem:deteminanat}, and equation (\ref{eq:eq7}) follows lemma \ref{lem:potential-bound}.

With the proof of the two propositions, we can conclude the Theorem \ref{thm:regret}, showing the regret bound as 
\begin{equation}
R_T \leq b \sigma \sqrt{T \left(d \log \left(1 + \frac{TB^2 W^2}{d \sigma^2} + \frac{\sum_{a=1}^A T^a (u_{t,k}^a)^2}{d \sigma^2}\right) + \log \left(\frac{4}{\delta}\right)\right)}.
\end{equation}
To prove the sub-linear regret bound, we need to analyze and simplify the dominant terms within the regret bound.
\paragraph{Dominant term analysis.}

To identify the dominant term, we carefully analyze how each term scales with \( T \):

\textbf{Term 1:} \(\frac{TB^2 W^2}{d \sigma^2}\), which grows linearly with \( T \).

\textbf{Term 2:} \(\frac{\sum_{a=1}^A T^a (u_{t,k}^a)^2}{d \sigma^2}\), which scales with \(\sum_{a=1}^A T^a\), which is at most \( T \), as not all arms may apply the {$k$}-NN adjustment at every time step. This sum represents an upper bound, capturing the maximum possible contribution from the {$k$}-NN component.
\paragraph{Simplifying the logarithmic term.}

Considering both terms inside the logarithm, we have
\begin{equation}
\log \left(1 + \frac{TB^2 W^2}{d \sigma^2} + \frac{\sum_{a=1}^A T^a (u_{t,k}^a)^2}{d \sigma^2}\right),
\end{equation}
which for large \( T \), we can approximate the logarithm as
\begin{equation}
\log \left(1 + \frac{TB^2 W^2}{d \sigma^2} + \frac{\sum_{a=1}^A T^a (u_{t,k}^a)^2}{d \sigma^2}\right) \approx \log \left(\frac{T (B^2 W^2 + \sum_{a=1}^A (u_{t,k}^a)^2)}{d \sigma^2}\right).
\end{equation}

\paragraph{Refined bound.}

Given that both terms grow with \( T \), for large \( T \), we have
\begin{equation}
\log \left(1 + \frac{T (B^2 W^2 + \sum_{a=1}^A (u_{t,k}^a)^2)}{d \sigma^2}\right).
\end{equation}

So, the regret bound becomes
\begin{equation}
R_T \leq b \sigma \sqrt{T \left(d \log \left(\frac{T (B^2 W^2 + \sum_{a=1}^A (u_{t,k}^a)^2)}{d \sigma^2}\right)\right)}.
\end{equation}
And for large \( T \), we have
\begin{equation}
R_T = O\left(\sigma \sqrt{dT \log T}\right).
\end{equation}

Without assuming any term is negligible, the regret of LNUCB-TA is optimal up to
\begin{equation}
R_T = O(\sqrt{dT \log T}).
\end{equation}
And by absorbing logarithmic factors into \(\tilde{O}\), we can state
\begin{equation}
R_T = \tilde{O}(\sqrt{dT}).
\end{equation}

This result establishes the optimality and efficiency of LNUCB-TA in achieving sub-linear regret, proving Theorem \ref{thm:regret}.

\subsubsection {Proof Theorem \ref{thm:exploration}}
We begin by considering the exploration parameter \( \alpha_{N_t^a} \), which is dynamically updated as:

\begin{equation}
\alpha_{N_t^a} = \frac{\alpha_0}{N_t^a + 1} \cdot \left( \kappa g + (1 - \kappa) n_t^a \right),
\end{equation}

where \( g \) represents the global attention and \( n_t^a \) is the local attention for arm \( a \) up to time \( t \). Specifically, the global attention \( g \) is defined as:

\begin{equation}
g = \frac{1}{A} \sum_{a=1}^{A} \overline{Y}^a,
\end{equation}

with \( A \) being the number of arms and \( \overline{Y}^a \) the average reward of arm \( a \). The local attention \( n_t^a \) is given by:

\begin{equation}
n_t^a = \frac{1}{N_t^a} \sum_{s=1}^{t-1} \hat{Y}_s^a,
\end{equation}

where \( N_t^a \) is the number of times arm \( a \) has been selected up to time \( t \), and \(\hat{Y}_s^a\) is the reward observed from arm \( a \) at time \( s \). Our goal is to compute \( \frac{d\alpha_{N_t^a}}{dN_t^a} \), representing the rate of change of the exploration parameter as \( N_t^a \) increases, i.e., how the system shifts from exploration to exploitation as more pulls are made on arm \( a \).

First, applying the product rule to differentiate \( \alpha_{N_t^a} \) with respect to \( N_t^a \), we have:

\begin{equation}
\frac{d\alpha_{N_t^a}}{dN_t^a} = \frac{d}{dN_t^a} \left( \frac{\alpha_0}{N_t^a + 1} \cdot \left( \kappa g + (1 - \kappa) n_t^a \right) \right).
\end{equation}

This can be expanded as:

\begin{equation}
\frac{d\alpha_{N_t^a}}{dN_t^a} = \frac{\alpha_0}{N_t^a + 1} \cdot \frac{d}{dN_t^a} \left( \kappa g + (1 - \kappa) n_t^a \right) + \left( \kappa g + (1 - \kappa) n_t^a \right) \cdot \frac{d}{dN_t^a} \left( \frac{\alpha_0}{N_t^a + 1} \right).
\end{equation}

Next, we compute the derivatives of each term separately. Since \( g \) is the global attention and does not depend on \( N_t^a \), its derivative is zero, and we only need to differentiate \( n_t^a \). Using the quotient rule, we compute the derivative of \( n_t^a \) as follows:

\begin{equation}
n_t^a = \frac{1}{N_t^a} \sum_{s=1}^{t-1} \hat{Y}_s^a,
\end{equation}

hence,

\begin{equation}
\frac{d n_t^a}{d N_t^a} = -\frac{1}{(N_t^a)^2} \sum_{s=1}^{t-1} \hat{Y}_s^a.
\end{equation}

Substituting this into the derivative of the first term:

\begin{equation}
\frac{d}{dN_t^a} \left( \kappa g + (1 - \kappa) n_t^a \right) = (1 - \kappa) \cdot \left( -\frac{1}{(N_t^a)^2} \sum_{s=1}^{t-1} \hat{Y}_s^a \right).
\end{equation}

For the second term, we differentiate \( \frac{\alpha_0}{N_t^a + 1} \) with respect to \( N_t^a \):

\begin{equation}
\frac{d}{dN_t^a} \left( \frac{\alpha_0}{N_t^a + 1} \right) = -\frac{\alpha_0}{(N_t^a + 1)^2}.
\end{equation}

Now, substituting these results back into the expression for \( \frac{d\alpha_{N_t^a}}{dN_t^a} \), we obtain:

\begin{equation}
\frac{d\alpha_{N_t^a}}{dN_t^a} = \frac{\alpha_0}{N_t^a + 1} \cdot (1 - \kappa) \cdot \left( -\frac{1}{(N_t^a)^2} \sum_{s=1}^{t-1} \hat{Y}_s^a \right) - \frac{\alpha_0}{(N_t^a + 1)^2} \cdot \left( \kappa g + (1 - \kappa) n_t^a \right).
\end{equation}

Expanding \( n_t^a \) in the second term gives:

\begin{equation}
n_t^a = \frac{1}{N_t^a} \sum_{s=1}^{t-1} \hat{Y}_s^a,
\end{equation}

so we substitute this into the second term to obtain:

\begin{equation}
\frac{d\alpha_{N_t^a}}{dN_t^a} = \frac{\alpha_0}{N_t^a + 1} \cdot (1 - \kappa) \cdot \left( -\frac{1}{(N_t^a)^2} \sum_{s=1}^{t-1} \hat{Y}_s^a \right) - \frac{\alpha_0}{(N_t^a + 1)^2} \cdot \left( \kappa g + (1 - \kappa) \cdot \frac{1}{N_t^a} \sum_{s=1}^{t-1} \hat{Y}_s^a \right).
\end{equation}

We can further expand both terms. The first term becomes:

\begin{equation}
\frac{\alpha_0}{N_t^a + 1} \cdot (1 - \kappa) \cdot \left( -\frac{1}{(N_t^a)^2} \sum_{s=1}^{t-1} \hat{Y}_s^a \right) = - \frac{\alpha_0 (1 - \kappa)}{(N_t^a + 1) \cdot (N_t^a)^2} \sum_{s=1}^{t-1} \hat{Y}_s^a.
\end{equation}

The second term expands as:

\begin{equation}
- \frac{\alpha_0}{(N_t^a + 1)^2} \cdot \left( \kappa g + (1 - \kappa) \cdot \frac{1}{N_t^a} \sum_{s=1}^{t-1} \hat{Y}_s^a \right).
\end{equation}

This can be split into two parts:

\begin{equation}
- \frac{\alpha_0 \kappa g}{(N_t^a + 1)^2} - \frac{\alpha_0 (1 - \kappa)}{(N_t^a + 1)^2 \cdot N_t^a} \sum_{s=1}^{t-1} \hat{Y}_s^a.
\end{equation}

Finally, the complete expanded expression for \( \frac{d\alpha_{N_t^a}}{dN_t^a} \) is:

\begin{equation}
\frac{d\alpha_{N_t^a}}{dN_t^a} = - \frac{\alpha_0 (1 - \kappa)}{(N_t^a + 1) \cdot (N_t^a)^2} \sum_{s=1}^{t-1} \hat{Y}_s^a - \frac{\alpha_0 \kappa g}{(N_t^a + 1)^2} - \frac{\alpha_0 (1 - \kappa)}{(N_t^a + 1)^2 \cdot N_t^a} \sum_{s=1}^{t-1} \hat{Y}_s^a.
\end{equation}

This result shows how the exploration parameter \( \alpha_{N_t^a} \) decreases as \( N_t^a \) increases, driven by both local attention (\( n_t^a \)) and global attention (\( g \)). The terms decay quadratically with \( N_t^a \), highlighting the shift from exploration to more focused exploitation as more observational data is gathered and the rewards from each arm become better understood.

\section {Additional Results}
\label{sec:more results}
\begin{figure}[!htb]
\begin{center}
	\fbox{\includegraphics[width=0.98\textwidth]{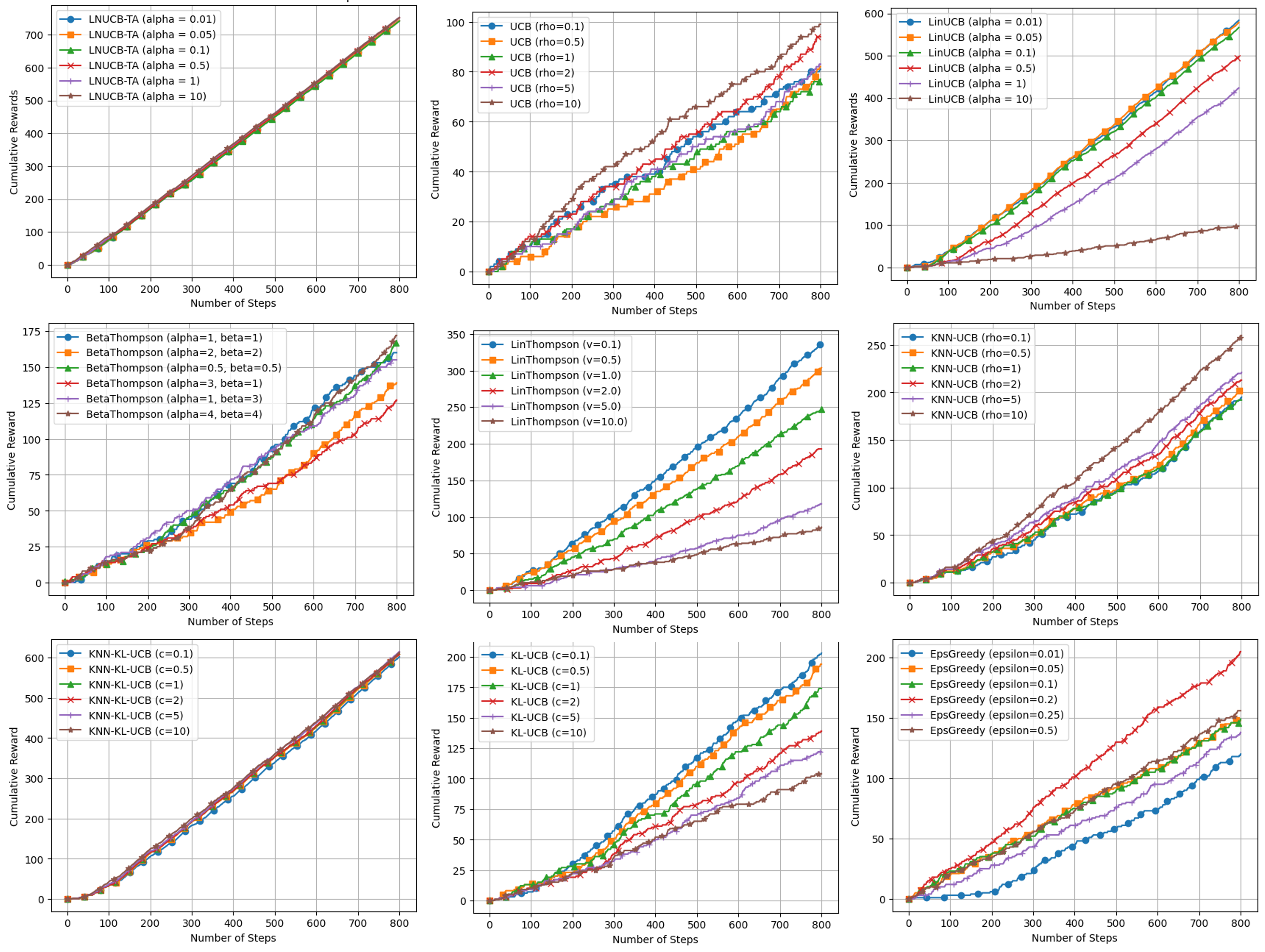}} 
\end{center}
\vspace{1em} 
\caption{Performance comparison of models based on cumulative reward across six distinct parameter settings. The LNUCB-TA model demonstrates superior performance and more stable results compared to other models.}
\label{fig:cumulative}
\end{figure}

\begin{figure*}[!htb]
\begin{center}
	\fbox{\includegraphics[width=0.98\textwidth]{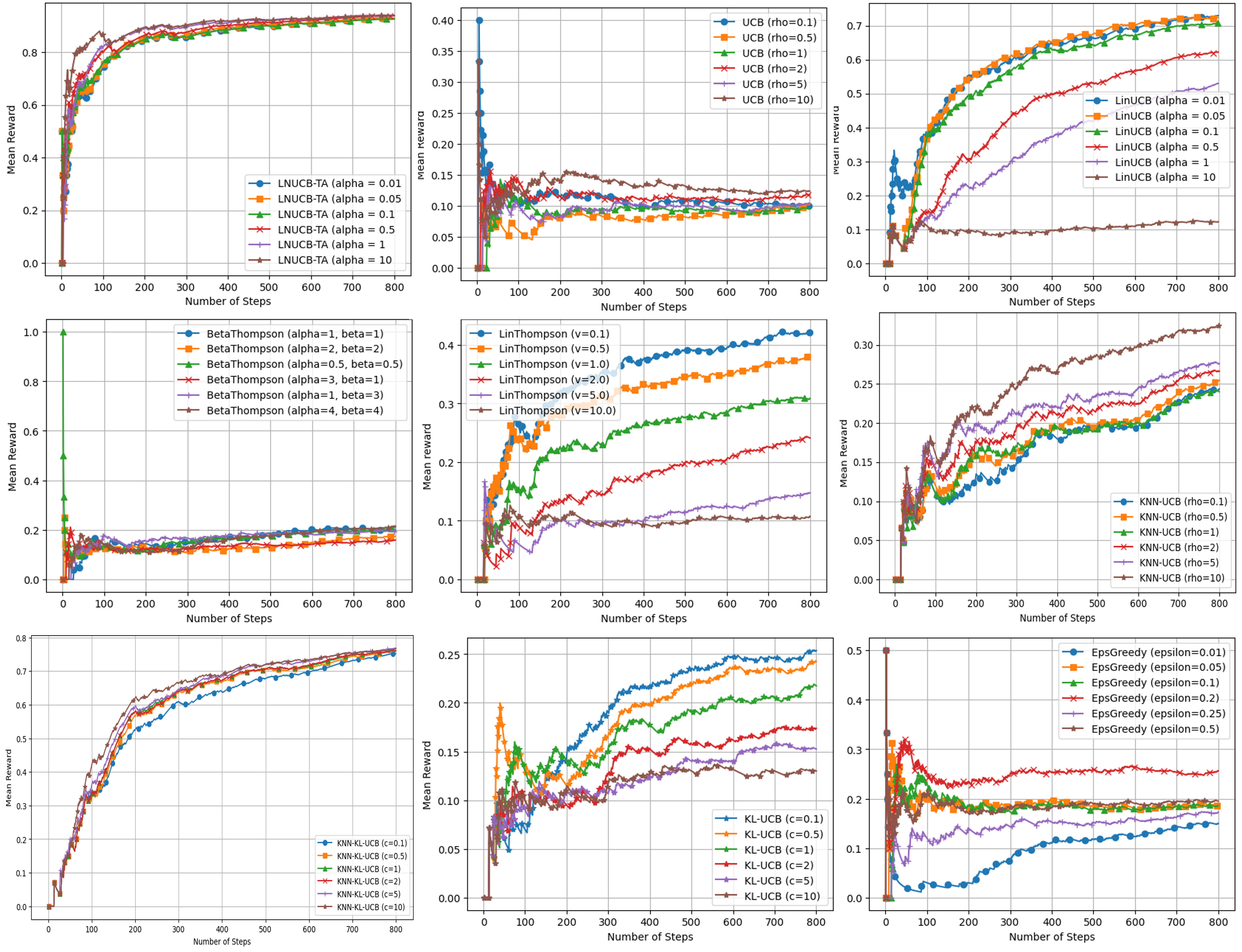}} 
\end{center}
\vspace{1em} 
\caption{Comparison of model performance based on mean reward across six distinct parameter settings. The LNUCB-TA model achieves the highest mean rewards and exhibits stable performance.}
\label{fig:mean}
\end{figure*}

In this Section, more quantitative results are provided.
\paragraph{Analysis of models with different parameters.} The experimental results, shown in Figures \ref{fig:cumulative} and \ref{fig:mean} and summarized in Table \ref{table:comparison-table}, highlight the performance of various MAB algorithms across different parameter settings. In this section, we set \(\kappa = 0.5\), \(\theta_{\text{min}} = 1\), and \(\theta_{\text{max}} = 5\). The maximum value of $k$ for $k$-NN KL-UCB and $k$-NN UCB is considered to be 5 to ensure a fair comparison among the models. The BetaThompson model, which was tested with six combinations of \((\alpha, \beta)\) parameters, achieved its best performance with parameters \((4, 4)\), resulting in a mean reward of 0.22 and a cumulative reward of 176. Similarly, the Epsilon Greedy algorithm, evaluated with six different \(\epsilon\) values, achieved the highest mean reward of 0.26 and a cumulative reward of 208 at \(\epsilon = 0.2\). KL-UCB, another prominent algorithm, demonstrated its best performance at \(c = 0.1\), with a mean reward of 0.25 and a cumulative reward of 200. {$k$}-NN KL-UCB and {$k$}-NN UCB, incorporating {$k$}-Nearest Neighbors, showed optimal results at \(c = 5\) and \(\rho = 10\), respectively, with mean rewards of 0.76 and 0.34. Notably, LinThompson and LinUCB algorithms, which leverage linear estimations, achieved mean rewards of 0.42 and 0.73, with cumulative rewards of 336 and 584. The UCB algorithm, when tested with six different \(\rho\) values, performed best at \(\rho = 10\), resulting in a mean reward of 0.14 and a cumulative reward of 112. 

As indicated in Table \ref{table:comparison-table}, our novel LNUCB-TA model significantly outperformed all the aforementioned algorithms, achieving a mean reward of 0.94 and a cumulative reward of 753. The improvement by LNUCB-TA over other models is substantial, with the highest relative improvement observed over UCB (572\%), followed by BetaThompson (327\%), Epsilon Greedy (262\%), KL-UCB (276\%), {$k$}-NN UCB (176\%), LinThompson (124\%), LinUCB (28\%), and {$k$}-NN KL-UCB (23\%). The significant enhancement and consistent performance underscore the robustness and effectiveness of the LNUCB-TA model, particularly its integration of linear and nonlinear estimations, adaptive {$k$}-Nearest Neighbors, and an attention-based exploration mechanism.
\begin{table}[!htb]
\centering
\begin{center}
	\begin{tabularx}{\textwidth}{l p{1.3cm} p{3cm} p{1.5cm} p{1.0cm} p{1.0cm} p{2cm}}
		\toprule
		\textbf{Model} & \textbf{Param.} & \textbf{Vals.} & \textbf{Best Param.} & \textbf{BMR} & \textbf{BCR} & \textbf{Imp. by LNUCB-TA (\%)} \\
		\midrule
		BetaThompson & \((\alpha, \beta)\) & (1, 1), (2, 2), (0.5, 0.5), (3, 1), (1, 3), (4, 4) & (4, 4) & 0.22 & 176 & 327.27 \\
		Epsilon Greedy & \(\epsilon\) & 0.01, 0.05, 0.1, 0.2, 0.25, 0.5 & 0.2 & 0.26 & 208 & 262.98 \\
		KL-UCB & \(c\) & 0.1, 0.5, 1, 2, 5, 10 & 0.1 & 0.25 & 200 & 276.50 \\
		{$k$}-NN KL-UCB & \(c\) & 0.1, 0.5, 1, 2, 5, 10 & 5 & 0.76 & 608 & 23.87 \\
		{$k$}-NN UCB & \(\rho\) & 0.1, 0.5, 1, 2, 5, 10 & 10 & 0.34 & 272 & 176.47 \\
		LinThompson & \(v\) & 0.1, 0.5, 1, 2, 5, 10 & 0.1 & 0.42 & 336 & 124.11 \\
		LinUCB & \(\alpha\) & 0.01, 0.05, 0.1, 0.5, 1, 10 & 0.01 & 0.73 & 584 & 28.91 \\
		UCB & \(\rho\) & 0.1, 0.5, 1, 2, 5, 10 & 10 & 0.14 & 112 & 572.32 \\
		LNUCB-TA & \(\alpha\) & 0.01, 0.05, 0.1, 0.5, 1, 10 & 1 & 0.94 & 753 & N/A \\ 
		\bottomrule
	\end{tabularx}
\end{center}

\caption{Comparison of model parameters and performance: The table summarizes the various models (Model), the parameters tested (Param.), their values (Vals.), and the best-performing parameters (Best Param.). It also includes the best mean reward (BMR) and best cumulative reward (BCR) achieved by each model, as well as the percentage improvement of our model LNUCB-TA compared to others (Imp. by LNUCB-TA).}
\label{table:comparison-table}
\end{table}

\begin{figure}[!htb]
\centering
\fbox{\includegraphics[width=0.98\textwidth]
	{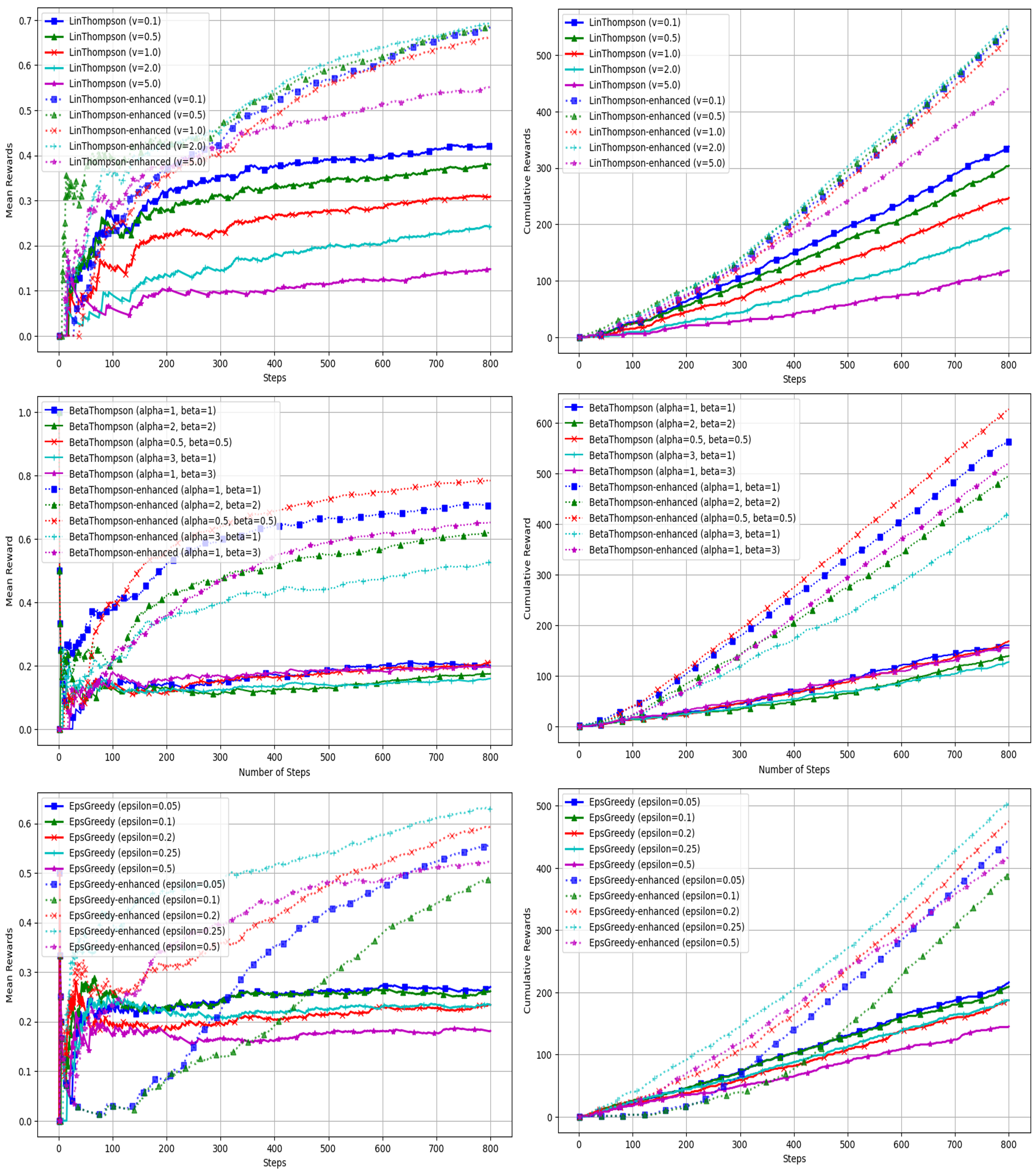}} 
\caption{Performance enhancements achieved by integrating the {$k$}-NN adaptive strategy in Algorithm \ref{alg:k-NN} and the attention mechanism in equation (\ref{eq:new-attention}) into traditional models BetaThompson, Epsilon Greedy, and LinThompson.}
\label{fig:enhanced}
\end{figure}

\paragraph{Improvement over other models.} Figure \ref{fig:enhanced} illustrates the performance enhancements achieved by integrating the {$k$}-NN adaptive strategy and an attention mechanism inspired by \citep{vaswani2017attention} (for each arm $a$ at time $t$, the exploration rate is weighted by an attention score as 
\begin{equation} 
\text{attention-score} = \frac{\exp(-\gamma \cdot N_t^a)}{\sum(\exp(-\gamma \cdot N_t^a))},\label{eq:new-attention}
\end{equation} where, $\gamma$ is a scaling parameter) into three traditional models namely BetaThompson, Epsilon Greedy, and LinThompson. Each enhanced model demonstrates a marked improvement in both cumulative and mean rewards over 800 steps. Specifically, the BetaThompson-enhanced model, with the best parameter combination \((\alpha, \beta) = (0.5, 0.5)\), achieves a mean reward of 0.79 and a cumulative reward of 632. Similarly, the Epsilon Greedy-enhanced model, optimized with \(\epsilon = 0.25\), reaches a mean reward of 0.58 and a cumulative reward of 464. The LinThompson-enhanced model, with \(v = 2\), shows a significant increase in performance, attaining a mean reward of 0.69 and a cumulative reward of 552.

Table \ref{table:comparison-table-enhanced} summarizes these results highlights the substantial improvements over their respective base models. The BetaThompson-enhanced model shows a 259.09\% improvement over the base model, the Epsilon Greedy-enhanced model shows a 123.08\% improvement, and the LinThompson-enhanced model demonstrates a 64.29\% enhancement. Despite these significant gains, the comparison to the LNUCB-TA model reveals that while these enhancements are substantial, they still fall short of the performance of LNUCB-TA, which achieves a mean reward of 0.94. Specifically, the BetaThompson-enhanced model performs 16.08\% worse than LNUCB-TA, Epsilon Greedy-enhanced 38.38\% worse, and LinThompson-enhanced 26.69\% worse. The superior performance of LNUCB-TA is attributed to its unique combination of both linear and nonlinear estimations. The results highlight the impact of the two key novelties—adaptive {$k$}-NN and attention mechanisms—setting a new framework for MAB algorithms through these innovative enhancements.

\begin{table}[!htb]
\centering
\begin{center}
	\begin{tabularx}{\textwidth}{l 
			>{\raggedright\arraybackslash}p{1.7cm} 
			>{\raggedright\arraybackslash}p{1.2cm} 
			>{\raggedright\arraybackslash}p{1.2cm} 
			>{\raggedright\arraybackslash}p{1.9cm} 
			>{\raggedright\arraybackslash}p{1.9cm}}
		\toprule
		\textbf{Model} & \textbf{Best Param.} & \textbf{BMR} & \textbf{BCR} & \textbf{Imp. Over Base Model (\%)} & \textbf{Comp. to LNUCB-TA (\%)} \\
		\midrule
		BetaThompson-enhanced & (0.5, 0.5) & 0.79 & 632 & 259.09 & -16.08 \\
		Epsilon Greedy-enhanced &  0.25 & 0.58 & 464 & 123.08 & -38.38 \\
		LinThompson-enhanced & 2 & 0.69 & 552 & 64.29 & -26.69 \\ 
		\bottomrule
	\end{tabularx}
\end{center}
\caption{Performance Comparison of Enhanced Models: The table presents the best parameters (Best Param.), best mean reward (BMR), and best cumulative reward (BCR), the improvement percentage over the base model, and the comparison percentage to LNUCB-TA (Comp. to LNUCB-TA (\%)).}
\label{table:comparison-table-enhanced}
\end{table}

\begin{figure}[!htb]
\centering
\fbox{\includegraphics[width=0.98\textwidth]
	{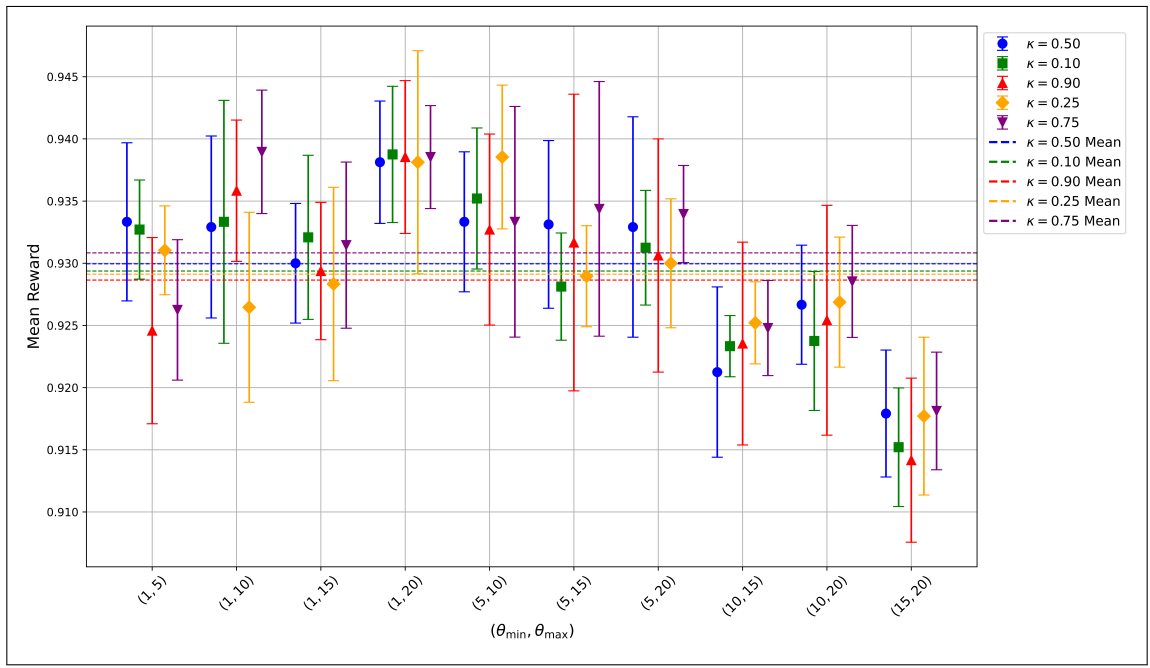}} 
\caption{Performance stability of LNUCB-TA across various parameter settings: The plot illustrates the mean reward ranges for different combinations of $\theta_{\mathrm{min}}$ and $\theta_{\mathrm{max}}$, and different values of $\kappa$. Despite variations in these parameters, the model consistently maintains high performance, underscoring its robustness and the effectiveness of integrating adaptive {$k$}-NN and attention mechanisms.}
\label{fig:errorbars}
\end{figure}

\paragraph{Error bars.} Based on the error bar plot in Figure ~\ref{fig:errorbars}, we can observe that the LNUCB-TA model demonstrates remarkable consistency in its performance across a variety of parameter settings. The plot shows the mean reward for different combinations of $\theta_{\text{min}}$ and $\theta_{\text{max}}$, and different values of $\kappa$, which is the weight of the global overall reward. Despite the changes in these parameters, the mean reward remains relatively stable, indicating that the model's performance is not heavily reliant on specific parameter choices. This consistency underscores the robustness of the LNUCB-TA model, making it a reliable choice for complex decision-making tasks where parameter tuning can be challenging.

\paragraph{Additional datasets.}
We extend our analysis of the LNUCB-TA model to additional real-world datasets to further validate its efficacy across diverse settings. One such dataset involves the AstroPh co-authorship network, initially observed at 5\% \citep{madhawa2019exploring}. Here, we focus on the cumulative reward comparison of our model against other state-of-the-art algorithms, demonstrating its capability in effectively expanding network visibility within a fixed query budget. Another dataset explored is an article matching dataset \citep{li2010contextual,li2011unbiased}, where the LNUCB-TA's performance is assessed in the context of matching relevant articles based on user preferences and interactions. These expanded evaluations provide a broader perspective on the model’s versatility and its applicability to complex, real-world problems such as network exploration and content recommendation.
\begin{figure}[!htb]
\begin{center}
	\fbox{\includegraphics[width=0.98\textwidth, height=0.40\textheight]{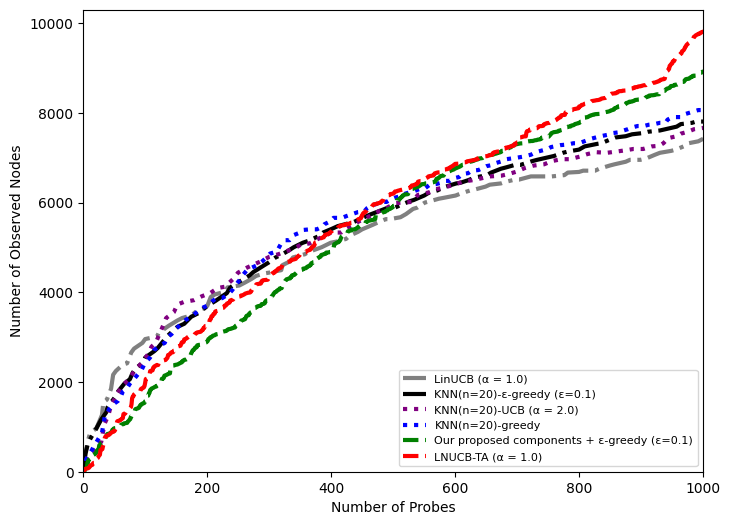}} 
\end{center}
\vspace{1em} 
\caption{Cumulative reward (y-axis) comparison of models on AstroPh co-authorship network initially observed at (5$\%$). Our \textbf{LNUCB-TA} model, represented by the \textbf{\textcolor{red}{red}} line, outperforms other models. Also, the \textbf{\textcolor{green}{green}} line, representing our novel {$k$}-NN approach with attention combined with $\epsilon$-Greedy, surpasses KNN-$\epsilon$-Greedy, showing the superiority of our proposed {$k$}-NN over existing {$k$}-NN bandit settings.}
\label{fig:network_dataset}
\end{figure}

In Figure \ref{fig:network_dataset}, the LNUCB-TA model, marked by the bold red line, outperforms other models with its superior performance as the evaluation progresses. This highlights the model's efficiency in adapting and optimizing its strategy over time, solidifying its effectiveness in dynamic settings. Additionally, our innovative approach that integrates {$k$}-NN with an attention mechanism into the $\epsilon$-Greedy strategy is represented by the bold green line. This combination shows significant improvements over the traditional KNN-$\epsilon$-Greedy model, underscoring the effectiveness of our proposed modifications in handling the exploration-exploitation balance more dynamically and efficiently. 

\begin{figure}[!htb]
\begin{center}
	\fbox{\includegraphics[width=0.98\textwidth, height=0.40\textheight]{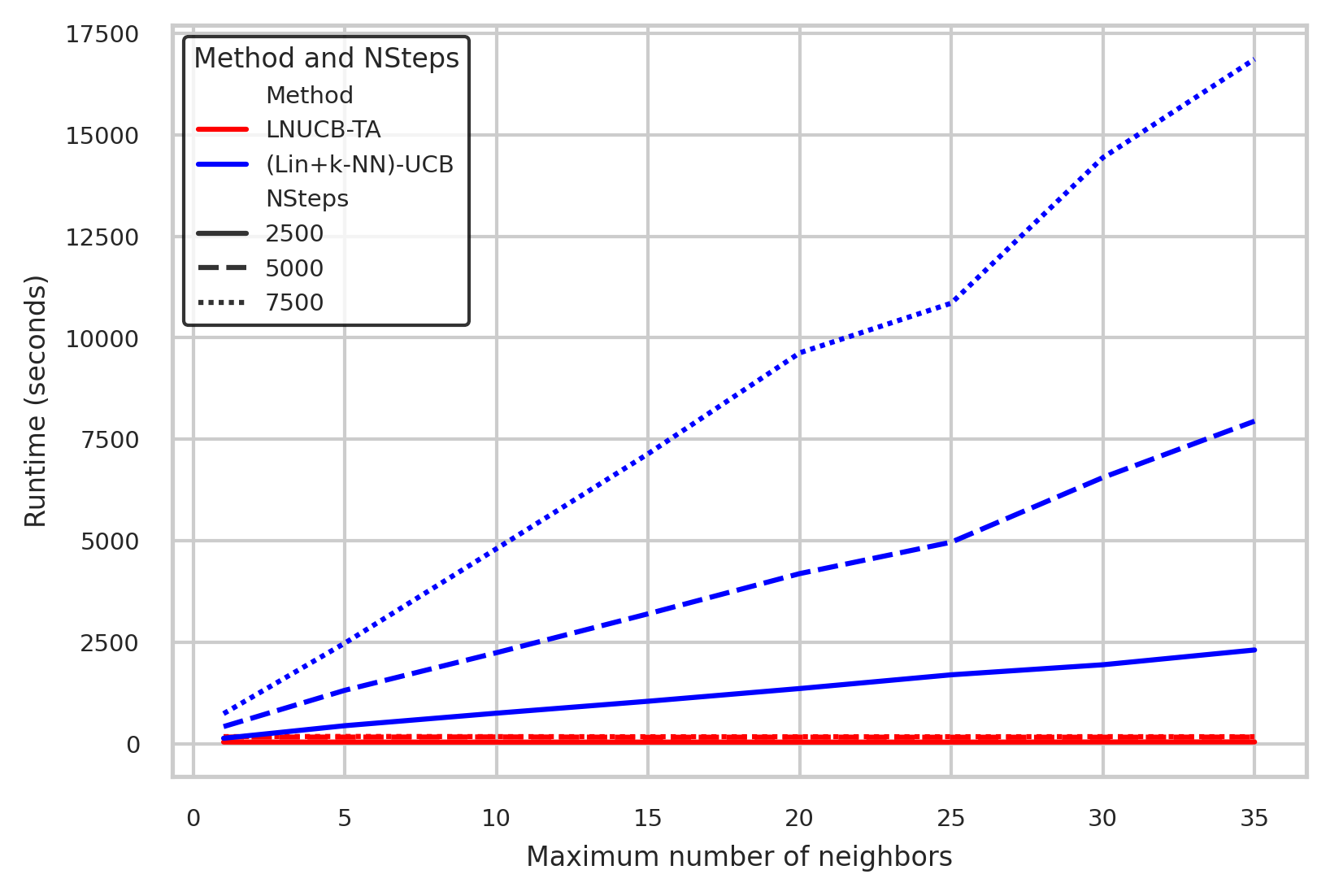}} 
\end{center}
\vspace{1em} 
\caption{Runtime and scalability comparison of our model against the straightforward combination model on the article matching dataset. The \textbf{LNUCB-TA} model is more scalable, maintaining quite consistent processing times, even as \(\max_k\) and the number of steps increase.}
\label{fig: runtime scalability}
\end{figure}

Figure \ref{fig: runtime scalability} presents the difference runtime between our proposed model against the vanilla combination of LinUCB and {$k$}-NN UCB model. The LNUCB-TA model, represented by the bold red line, consistently exhibits the lowest runtime, particularly as the maximum number of neighbors increases, underscoring its computational efficiency compared to the Lin+{$k$}-NN-UCB model (blue line) and other setups denoted by the dotted lines for varying NSteps. This demonstrates the LNUCB-TA model's capability to maintain lower computational costs even as the complexity of the task increases.

Additionally, the results presented in Table \ref{table:article_matching} shows that the LNUCB-TA model consistently outperforms purely linear models, purely nonlinear models, and the vanilla combination of linear and nonlinear approaches in terms of cumulative rewards across various exploration rates and operational steps. For instance, at an exploration rate of 0.1 and 7500 steps, it achieves the highest cumulative reward of 7261. The model is also substantially more efficient than the straightforward combination model (Lin+{$k$}-NN)-UCB, which takes 3381.71 seconds for a lower reward score, compared to the LNUCB-TA's 102.00 seconds.

\begin{table}[!htb]
\begin{center}
	\begin{tabular}{@{}>{\centering\arraybackslash}p{0.5cm} 
			>{\centering\arraybackslash}p{0.7cm} 
			>{\centering\arraybackslash}p{1.1cm} 
			>{\centering\arraybackslash}p{1.2cm} 
			>{\centering\arraybackslash}p{1.0cm} 
			>{\centering\arraybackslash}p{1.0cm} 
			>{\centering\arraybackslash}p{1.2cm} 
			>{\centering\arraybackslash}p{1.2cm} 
			>{\centering\arraybackslash}p{1.5cm} 
			>{\centering\arraybackslash}p{1.4cm}@{}}
		\toprule
		$\alpha/\rho$ & Steps & LinUCB (CR) & LinUCB Runtime & {$k$}-NN UCB (CR) & {$k$}-NN UCB Runtime & (Lin+{$k$}-NN)-UCB (CR) & (Lin+{$k$}-NN)-UCB Runtime & \textbf{LNUCB-TA} (CR) & LNUCB-TA Runtime \\ \midrule
		0.1          & 2500  & 2089        & 10.11          & 1618           & 14.79             & 2126               & 287.85                 & \textbf{2262}         & 26.21          \\
		0.1          & 5000  & 4570        & 12.36          & 3763           & 35.80             & 4604               & 1333.2                 & \textbf{4762}         & 92.01          \\
		0.1          & 7500  & 7063        & 19.79          & 6004           & 62.92             & 7099               & 3381.71                & \textbf{7261}         & 102.00         \\
		1            & 2500  & 1349        & 5.80           & 1607           & 15.98             & 1401               & 295.59                 & \textbf{1997}         & 24.08          \\
		1            & 5000  & 3720        & 12.30          & 3739           & 36.17             & 3785               & 1331.09                & \textbf{4497}         & 58.43          \\
		1            & 7500  & 6149        & 19.02          & 5996           & 62.03             & 6186               & 3226.52                & \textbf{6996}         & 98.57          \\
		10           & 2500  & 410         & 6.53           & 1595           & 15.84             & 410                & 279.34                 & \textbf{1601}         & 21.65          \\
		10           & 5000  & 1197        & 13.37          & 3721           & 36.57             & 1311               & 1169.57                & \textbf{4019}         & 55.50          \\
		10           & 7500  & 2282        & 18.07          & 5966           & 61.90             & 2536               & 3223.6                 & \textbf{6519}         & 95.48          \\ \bottomrule
	\end{tabular}
\end{center}
\centering
\caption{Comparison of models on the article matching dataset, using a maximum of 5 neighbors based on cumulative reward (CR). We observe varying performance between the purely linear, purely nonlinear, and the vanilla combination model with neither of them demonstrating absolute dominance. However, the \textbf{LNUCB-TA} model consistently outperforms all three of them.}
\label{table:article_matching}
\end{table}

\section {Limitation and future direction}
\label{sec:limitation}
\paragraph{Limitation.} 
One limitation of our approach is the assumption of equal weights for the linear and nonlinear components in the model. While this simplifies the model, it may not fully capture the complexities of the underlying data. Future work could explore assigning different weights to these components, potentially enhancing performance by better capturing the data's structure. Additionally, the weights could be dynamically adjusted for each arm at each time step using attention mechanisms, further improving adaptability.

Also, our current implementation of the GALA mechanism employs a fixed weight (\(\kappa\)) to balance global and local attention in adjusting the exploration factor. While we have tried different fixed values for \(\kappa\), these weights might be not optimized. Determining the optimal value for \(\kappa\) could further enhance the model's performance.

\paragraph{Attention mechanisms in MAB frameworks.} The introduction of attention mechanisms in the MAB framework opens new avenues for enhancing decision-making processes in various domains. While our work applied attention to the exploration rate, there are numerous other areas within the MAB framework where attention mechanisms can be beneficial. For instance, attention could be used to dynamically prioritize contexts based on their significance or complexity, thereby improving overall efficiency and effectiveness. Additionally, attention mechanisms could be applied to weight the influence of historical rewards differently over time, allowing for more nuanced learning from past experiences. Another potential application could be the use of attention to identify and focus on emerging trends or shifts in the data, ensuring that the model adapts swiftly to new patterns
\paragraph{Impact on industrial settings.}
LNUCB-TA, as it dynamically adjusts its exploration rate, can be beneficial in various areas where initial parameters need to be optimized, such as in recommendation systems \citep{zhou2017large,bouneffouf2012contextual,bouneffouf2014contextual} where initial user preferences are unknown, in finance \citep{shen2015portfolio,huo2017risk} for portfolio optimization where initial risk preferences must be set, and in healthcare \citep{bastani2020online,durand2018contextual} for personalized treatment plans where patient-specific parameters need to be optimized.
Our model can also be applied to areas not yet extensively covered by MAB approaches \citep{bouneffouf2019survey}, such as manufacturing. In this context, each arm represents a different material or material property configuration, while the context includes features describing the manufacturing conditions and requirements. The reward corresponds to the performance or suitability of the material under these conditions. By leveraging both linear and nonlinear estimations along with attention-based mechanisms, LNUCB-TA can effectively balance exploration and exploitation, identifying optimal material properties under varying conditions. This ability to dynamically adapt and refine decisions based on historical data and contextual insights makes our model particularly well-suited for such applications.

\paragraph{A new paradigm for MAB algorithms.}
Moreover, the incorporation of adaptive {$k$}-NN discussed in Algorithm \ref{alg:k-NN}, and attention mechanisms discussed in Algorithm \ref{alg:attention} and equation (\ref{eq:new-attention}) not only enhances the performance of LNUCB-TA but also improves the performance of other models. This sets a new framework for MAB algorithms by integrating these advanced modifications. 

\paragraph{Technical extensions in other areas.}
The inspiration from how we used attention mechanisms to make our model independent of initial parameter choices can be applied in various technical fields. This approach can enhance meta-heuristic algorithms for combinatorial optimization problems \citep{agushaka2022initialisation,shadkam2022parameter}, evolutionary algorithms where initial population parameters must be set \citep{lobo2007parameter,qin2023self}, and machine and federated learning models where hyperparameters need to be tuned \citep{koskela2024practical,khodak2021federated,turner2021bayesian}. By reducing the dependency on the initial parameter settings, this concept can improve the robustness and efficiency of these techniques, ensuring consistent performance irrespective of the chosen initial parameters.

\end{document}

%% file: math_commands.tex
\usepackage{amsmath,amsfonts,bm}

\def\eqref#1{equation~\ref{#1}}

\def\1{\bm{1}}

\DeclareMathAlphabet{\mathsfit}{\encodingdefault}{\sfdefault}{m}{sl}
\SetMathAlphabet{\mathsfit}{bold}{\encodingdefault}{\sfdefault}{bx}{n}













%% file: LNUCB-TA_JMLR.bbl
\begin{thebibliography}{83}
\providecommand{\natexlab}[1]{#1}
\providecommand{\url}[1]{\texttt{#1}}
\expandafter\ifx\csname urlstyle\endcsname\relax
  \providecommand{\doi}[1]{doi: #1}\else
  \providecommand{\doi}{doi: \begingroup \urlstyle{rm}\Url}\fi

\bibitem[Abbasi-Yadkori et~al.(2011)Abbasi-Yadkori, P{\'a}l, and
  Szepesv{\'a}ri]{abbasi2011improved}
Yasin Abbasi-Yadkori, D{\'a}vid P{\'a}l, and Csaba Szepesv{\'a}ri.
\newblock Improved algorithms for linear stochastic bandits.
\newblock \emph{Advances in Neural Information Processing Systems}, 24, 2011.

\bibitem[Agarwal et~al.(2019)Agarwal, Jiang, Kakade, and
  Sun]{agarwal2019reinforcement}
Alekh Agarwal, Nan Jiang, Sham~M. Kakade, and Wen Sun.
\newblock Reinforcement learning: Theory and algorithms.
\newblock \emph{CS Dept., UW Seattle, Seattle, WA, USA, Tech. Rep},
  32:\penalty0 96, 2019.

\bibitem[Agrawal and Goyal(2013)]{agrawal2013thompson}
Shipra Agrawal and Navin Goyal.
\newblock Thompson sampling for contextual bandits with linear payoffs.
\newblock In \emph{International Conference on Machine Learning}, pages
  127--135. PMLR, 2013.

\bibitem[Agushaka and Ezugwu(2022)]{agushaka2022initialisation}
Jeffrey~O. Agushaka and Absalom~E. Ezugwu.
\newblock Initialisation approaches for population-based metaheuristic
  algorithms: A comprehensive review.
\newblock \emph{Applied Sciences}, 12\penalty0 (2):\penalty0 896, 2022.

\bibitem[Allesiardo et~al.(2014)Allesiardo, F{\'e}raud, and
  Bouneffouf]{allesiardo2014neural}
Robin Allesiardo, Rapha{\"e}l F{\'e}raud, and Djallel Bouneffouf.
\newblock A neural networks committee for the contextual bandit problem.
\newblock In \emph{Neural Information Processing: 21st International
  Conference, {ICONIP} 2014}, volume 8834 of \emph{Lecture Notes in Computer
  Science}, pages 1--21. Springer International Publishing, 2014.

\bibitem[Alon et~al.(2015)Alon, Cesa-Bianchi, Dekel, and Koren]{alon2015online}
Noga Alon, Nicolo Cesa-Bianchi, Ofer Dekel, and Tomer Koren.
\newblock Online learning with feedback graphs: Beyond bandits.
\newblock In \emph{Conference on Learning Theory}, pages 23--35. PMLR, 2015.

\bibitem[Asuncion et~al.(2007)Asuncion, Newman, et~al.]{asuncion2007uci}
Arthur Asuncion, David Newman, et~al.
\newblock Uci machine learning repository, 2007.

\bibitem[Audibert et~al.(2009)Audibert, Munos, and
  Szepesv{\'a}ri]{audibert2009exploration}
Jean-Yves Audibert, R{\'e}mi Munos, and Csaba Szepesv{\'a}ri.
\newblock Exploration--exploitation tradeoff using variance estimates in
  multi-armed bandits.
\newblock \emph{Theoretical Computer Science}, 410\penalty0 (19):\penalty0
  1876--1902, 2009.

\bibitem[Auer et~al.(2002{\natexlab{a}})Auer, Cesa-Bianchi, and
  Fischer]{auer2002finite}
Peter Auer, Nicolo Cesa-Bianchi, and Paul Fischer.
\newblock Finite-time analysis of the multiarmed bandit problem.
\newblock \emph{Machine Learning}, 47:\penalty0 235--256, 2002{\natexlab{a}}.

\bibitem[Auer et~al.(2002{\natexlab{b}})Auer, Cesa-Bianchi, Freund, and
  Schapire]{auer2002nonstochastic}
Peter Auer, Nicolo Cesa-Bianchi, Yoav Freund, and Robert~E. Schapire.
\newblock The nonstochastic multiarmed bandit problem.
\newblock \emph{SIAM Journal on Computing}, 32\penalty0 (1):\penalty0 48--77,
  2002{\natexlab{b}}.

\bibitem[Axler(2015)]{axler2015linear}
Sheldon Axler.
\newblock \emph{Linear Algebra Done Right}.
\newblock Springer, 2015.

\bibitem[Aziz et~al.(2021)Aziz, Kaufmann, and Riviere]{aziz2021multi}
Maryam Aziz, Emilie Kaufmann, and Marie-Karelle Riviere.
\newblock On multi-armed bandit designs for dose-finding trials.
\newblock \emph{Journal of Machine Learning Research}, 22\penalty0
  (14):\penalty0 1--38, 2021.

\bibitem[Bastani and Bayati(2020)]{bastani2020online}
Hamsa Bastani and Mohsen Bayati.
\newblock Online decision making with high-dimensional covariates.
\newblock \emph{Operations Research}, 68\penalty0 (1):\penalty0 276--294, 2020.

\bibitem[Boissonnat et~al.(2018)Boissonnat, Chazal, and
  Yvinec]{boissonnat2018geometric}
Jean-Daniel Boissonnat, Fr{\'e}d{\'e}ric Chazal, and Mariette Yvinec.
\newblock \emph{Geometric and Topological Inference}, volume~57.
\newblock Cambridge University Press, 2018.

\bibitem[Bouneffouf and Rish(2019)]{bouneffouf2019survey}
Djallel Bouneffouf and Irina Rish.
\newblock A survey on practical applications of multi-armed and contextual
  bandits.
\newblock \emph{arXiv preprint arXiv:1904.10040}, 2019.

\bibitem[Bouneffouf et~al.(2012)Bouneffouf, Bouzeghoub, and
  Gan{\c{c}}arski]{bouneffouf2012contextual}
Djallel Bouneffouf, Amel Bouzeghoub, and Alda~Lopes Gan{\c{c}}arski.
\newblock A contextual-bandit algorithm for mobile context-aware recommender
  system.
\newblock In \emph{Neural Information Processing: 19th International
  Conference, ICONIP 2012, Doha, Qatar, November 12-15, 2012, Proceedings, Part
  III 19}, pages 324--331. Springer, 2012.

\bibitem[Bouneffouf et~al.(2014)Bouneffouf, Laroche, Urvoy, F{\'e}raud, and
  Allesiardo]{bouneffouf2014contextual}
Djallel Bouneffouf, Romain Laroche, Tanguy Urvoy, Raphael F{\'e}raud, and Robin
  Allesiardo.
\newblock Contextual bandit for active learning: Active thompson sampling.
\newblock In \emph{Neural Information Processing: 21st International
  Conference, ICONIP 2014, Kuching, Malaysia, November 3-6, 2014. Proceedings,
  Part I 21}, pages 405--412. Springer, 2014.

\bibitem[Bouneffouf et~al.(2020)Bouneffouf, Rish, and
  Aggarwal]{bouneffouf2020survey}
Djallel Bouneffouf, Irina Rish, and Charu Aggarwal.
\newblock Survey on applications of multi-armed and contextual bandits.
\newblock In \emph{2020 IEEE Congress on Evolutionary Computation (CEC)}, pages
  1--8. IEEE, 2020.

\bibitem[Bubeck et~al.(2012)Bubeck, Cesa-Bianchi, et~al.]{bubeck2012regret}
S{\'e}bastien Bubeck, Nicolo Cesa-Bianchi, et~al.
\newblock Regret analysis of stochastic and nonstochastic multi-armed bandit
  problems.
\newblock \emph{Foundations and Trends{\textregistered} in Machine Learning},
  5\penalty0 (1):\penalty0 1--122, 2012.

\bibitem[Carlsson et~al.(2021)Carlsson, Dubhashi, and
  Johansson]{carlsson2021thompson}
Emil Carlsson, Devdatt Dubhashi, and Fredrik~D. Johansson.
\newblock Thompson sampling for bandits with clustered arms.
\newblock \emph{arXiv preprint arXiv:2109.01656}, 2021.

\bibitem[Chowdhury and Gopalan(2017)]{chowdhury2017kernelized}
Sayak~Ray Chowdhury and Aditya Gopalan.
\newblock On kernelized multi-armed bandits.
\newblock In \emph{International Conference on Machine Learning}, pages
  844--853. PMLR, 2017.

\bibitem[Chu et~al.(2011)Chu, Li, Reyzin, and Schapire]{chu2011contextual}
Wei Chu, Lihong Li, Lev Reyzin, and Robert Schapire.
\newblock Contextual bandits with linear payoff functions.
\newblock In \emph{Proceedings of the Fourteenth International Conference on
  Artificial Intelligence and Statistics}, pages 208--214. JMLR Workshop and
  Conference Proceedings, 2011.

\bibitem[Curt{\`o} et~al.(2023)Curt{\`o}, de~Zarz{\`a}, Roig, Cano, Manzoni,
  and Calafate]{de2023llm}
J.~De Curt{\`o}, Irene de~Zarz{\`a}, Gemma Roig, Juan~Carlos Cano, Pietro
  Manzoni, and Carlos~T. Calafate.
\newblock Llm-informed multi-armed bandit strategies for non-stationary
  environments.
\newblock \emph{Electronics}, 12\penalty0 (13):\penalty0 2814, 2023.

\bibitem[Cvetkovski(2012)]{cvetkovski2012inequalities}
Zdravko Cvetkovski.
\newblock \emph{Inequalities: Theorems, Techniques and Selected Problems}.
\newblock Springer Science \& Business Media, 2012.

\bibitem[Dimakopoulou et~al.(2019)Dimakopoulou, Zhou, Athey, and
  Imbens]{dimakopoulou2019balanced}
Maria Dimakopoulou, Zhengyuan Zhou, Susan Athey, and Guido Imbens.
\newblock Balanced linear contextual bandits.
\newblock In \emph{Proceedings of the AAAI Conference on Artificial
  Intelligence}, volume~33, pages 3445--3453, 2019.

\bibitem[Ding et~al.(2021)Ding, Liu, Miao, Cheng, and Tang]{ding2021hybrid}
Qinxu Ding, Yong Liu, Chunyan Miao, Fei Cheng, and Haihong Tang.
\newblock A hybrid bandit framework for diversified recommendation.
\newblock In \emph{Proceedings of the AAAI Conference on Artificial
  Intelligence}, volume~35, pages 4036--4044, 2021.

\bibitem[Dong et~al.(2021)Dong, Yang, and Ma]{dong2021provable}
Kefan Dong, Jiaqi Yang, and Tengyu Ma.
\newblock Provable model-based nonlinear bandit and reinforcement learning:
  Shelve optimism, embrace virtual curvature.
\newblock \emph{Advances in Neural Information Processing Systems},
  34:\penalty0 26168--26182, 2021.

\bibitem[Durand et~al.(2018)Durand, Achilleos, Iacovides, Strati, Mitsis, and
  Pineau]{durand2018contextual}
Audrey Durand, Charis Achilleos, Demetris Iacovides, Katerina Strati,
  Georgios~D. Mitsis, and Joelle Pineau.
\newblock Contextual bandits for adapting treatment in a mouse model of de novo
  carcinogenesis.
\newblock In \emph{Machine Learning for Healthcare Conference}, pages 67--82.
  PMLR, 2018.

\bibitem[Eftekhari and Wakin(2015)]{eftekhari2015new}
Armin Eftekhari and Michael~B. Wakin.
\newblock New analysis of manifold embeddings and signal recovery from
  compressive measurements.
\newblock \emph{Applied and Computational Harmonic Analysis}, 39\penalty0
  (1):\penalty0 67--109, 2015.

\bibitem[Eleftheriadis et~al.(2024)Eleftheriadis, Evangelidis, and
  Ougiaroglou]{eleftheriadis2024empirical}
Stylianos Eleftheriadis, Georgios Evangelidis, and Stefanos Ougiaroglou.
\newblock An empirical analysis of data reduction techniques for k-nn
  classification.
\newblock In \emph{IFIP International Conference on Artificial Intelligence
  Applications and Innovations}, pages 83--97. Springer, 2024.

\bibitem[Federer(1959)]{federer1959curvature}
Herbert Federer.
\newblock Curvature measures.
\newblock \emph{Transactions of the American Mathematical Society}, 93\penalty0
  (3):\penalty0 418--491, 1959.

\bibitem[Folland(1999)]{folland1999real}
Gerald~B. Folland.
\newblock \emph{Real Analysis: Modern Techniques and Their Applications},
  volume~40.
\newblock John Wiley \& Sons, 1999.

\bibitem[Garivier and Capp{\'e}(2011)]{garivier2011kl}
Aur{\'e}lien Garivier and Olivier Capp{\'e}.
\newblock The kl-ucb algorithm for bounded stochastic bandits and beyond.
\newblock In \emph{Proceedings of the 24th Annual Conference on Learning
  Theory}, pages 359--376. JMLR Workshop and Conference Proceedings, 2011.

\bibitem[Goktas et~al.(2024)Goktas, Greenwald, Zhao, Koppel, and
  Ganesh]{goktas2024efficient}
Denizalp Goktas, Amy Greenwald, Sadie Zhao, Alec Koppel, and Sumitra Ganesh.
\newblock Efficient inverse multiagent learning.
\newblock In \emph{The Twelfth International Conference on Learning
  Representations}, 2024.

\bibitem[Harville(1998)]{harville1998matrix}
David~A. Harville.
\newblock Matrix algebra from a statistician's perspective, 1998.

\bibitem[Hillel et~al.(2013)Hillel, Karnin, Koren, Lempel, and
  Somekh]{hillel2013distributed}
Eshcar Hillel, Zohar~S. Karnin, Tomer Koren, Ronny Lempel, and Oren Somekh.
\newblock Distributed exploration in multi-armed bandits.
\newblock \emph{Advances in Neural Information Processing Systems}, 26, 2013.

\bibitem[Huo and Fu(2017)]{huo2017risk}
Xiaoguang Huo and Feng Fu.
\newblock Risk-aware multi-armed bandit problem with application to portfolio
  selection.
\newblock \emph{Royal Society Open Science}, 4\penalty0 (11):\penalty0 171377,
  2017.

\bibitem[Jia et~al.(2022)Jia, Zhang, Zhou, Gu, and Wang]{jia2022learning}
Yiling Jia, Weitong Zhang, Dongruo Zhou, Quanquan Gu, and Hongning Wang.
\newblock Learning neural contextual bandits through perturbed rewards.
\newblock \emph{arXiv preprint arXiv:2201.09910}, 2022.

\bibitem[Kamiura and Sano(2017)]{kamiura2017optimism}
Moto Kamiura and Kohei Sano.
\newblock Optimism in the face of uncertainty supported by a
  statistically-designed multi-armed bandit algorithm.
\newblock \emph{Biosystems}, 160:\penalty0 25--32, 2017.

\bibitem[Kassraie and Krause(2022)]{kassraie2022neural}
Parnian Kassraie and Andreas Krause.
\newblock Neural contextual bandits without regret.
\newblock In \emph{Proceedings of The 25th International Conference on
  Artificial Intelligence and Statistics}, pages 240--278. PMLR, 2022.

\bibitem[Khodak et~al.(2021)Khodak, Tu, Li, Li, Balcan, Smith, and
  Talwalkar]{khodak2021federated}
Mikhail Khodak, Renbo Tu, Tian Li, Liam Li, Maria-Florina~F. Balcan, Virginia
  Smith, and Ameet Talwalkar.
\newblock Federated hyperparameter tuning: Challenges, baselines, and
  connections to weight-sharing.
\newblock \emph{Advances in Neural Information Processing Systems},
  34:\penalty0 19184--19197, 2021.

\bibitem[Koskela and Kulkarni(2024)]{koskela2024practical}
Antti Koskela and Tejas~D. Kulkarni.
\newblock Practical differentially private hyperparameter tuning with
  subsampling.
\newblock \emph{Advances in Neural Information Processing Systems}, 36, 2024.

\bibitem[Lattimore and Szepesv{\'a}ri(2020)]{lattimore2020bandit}
Tor Lattimore and Csaba Szepesv{\'a}ri.
\newblock \emph{Bandit Algorithms}.
\newblock Cambridge University Press, 2020.

\bibitem[Lee(2006)]{lee2006riemannian}
John~M. Lee.
\newblock \emph{Riemannian Manifolds: An Introduction to Curvature}, volume
  176.
\newblock Springer Science \& Business Media, 2006.

\bibitem[Lee et~al.(2024)Lee, Park, and Lee]{lee2024soft}
Seunghan Lee, Taeyoung Park, and Kibok Lee.
\newblock Soft contrastive learning for time series.
\newblock In \emph{The Twelfth International Conference on Learning
  Representations}, 2024.

\bibitem[Li et~al.(2010)Li, Chu, Langford, and Schapire]{li2010contextual}
Lihong Li, Wei Chu, John Langford, and Robert~E. Schapire.
\newblock A contextual-bandit approach to personalized news article
  recommendation.
\newblock In \emph{Proceedings of the 19th International Conference on World
  Wide Web}, pages 661--670, 2010.

\bibitem[Li et~al.(2011)Li, Chu, Langford, and Wang]{li2011unbiased}
Lihong Li, Wei Chu, John Langford, and Xuanhui Wang.
\newblock Unbiased offline evaluation of contextual-bandit-based news article
  recommendation algorithms.
\newblock In \emph{Proceedings of the Fourth ACM International Conference on
  Web Search and Data Mining}, pages 297--306, 2011.

\bibitem[Linsley et~al.(2018)Linsley, Shiebler, Eberhardt, and
  Serre]{linsley2018learning}
Drew Linsley, Dan Shiebler, Sven Eberhardt, and Thomas Serre.
\newblock Learning what and where to attend.
\newblock \emph{arXiv preprint arXiv:1805.08819}, 2018.

\bibitem[Liu et~al.(2024{\natexlab{a}})Liu, Wang, Zheng, Hao, Bai, Ye, Wang,
  Piao, and Sun]{liu2024ovd}
Jinyi Liu, Zhi Wang, Yan Zheng, Jianye Hao, Chenjia Bai, Junjie Ye, Zhen Wang,
  Haiyin Piao, and Yang Sun.
\newblock Ovd-explorer: Optimism should not be the sole pursuit of exploration
  in noisy environments.
\newblock In \emph{Proceedings of the AAAI Conference on Artificial
  Intelligence}, volume~38, pages 13954--13962, 2024{\natexlab{a}}.

\bibitem[Liu et~al.(2024{\natexlab{b}})Liu, Zuo, Wang, Wang, Xu, and
  Lui]{liu2024learning}
Xutong Liu, Jinhang Zuo, Junkai Wang, Zhiyong Wang, Yuedong Xu, and John~CS
  Lui.
\newblock Learning context-aware probabilistic maximum coverage bandits: A
  variance-adaptive approach.
\newblock In \emph{IEEE INFOCOM 2024-IEEE Conference on Computer
  Communications}, pages 2189--2198. IEEE, 2024{\natexlab{b}}.

\bibitem[Lobo et~al.(2007)Lobo, Lima, and Michalewicz]{lobo2007parameter}
FJ~Lobo, Cl{\'a}udio~F. Lima, and Zbigniew Michalewicz.
\newblock \emph{Parameter Setting in Evolutionary Algorithms}, volume~54.
\newblock Springer Science \& Business Media, 2007.

\bibitem[Lykouris et~al.(2021)Lykouris, Simchowitz, Slivkins, and
  Sun]{lykouris2021corruption}
Thodoris Lykouris, Max Simchowitz, Alex Slivkins, and Wen Sun.
\newblock Corruption-robust exploration in episodic reinforcement learning.
\newblock In \emph{Conference on Learning Theory}, pages 3242--3245. PMLR,
  2021.

\bibitem[Madhawa and Murata(2019{\natexlab{a}})]{madhawa2019exploring}
Kaushalya Madhawa and Tsuyoshi Murata.
\newblock Exploring partially observed networks with nonparametric bandits.
\newblock In \emph{Complex Networks and Their Applications VII: Volume 2
  Proceedings The 7th International Conference on Complex Networks and Their
  Applications COMPLEX NETWORKS 2018 7}, pages 158--168. Springer,
  2019{\natexlab{a}}.

\bibitem[Madhawa and Murata(2019{\natexlab{b}})]{madhawa2019multi}
Kaushalya Madhawa and Tsuyoshi Murata.
\newblock A multi-armed bandit approach for exploring partially observed
  networks.
\newblock \emph{Applied Network Science}, 4:\penalty0 1--18,
  2019{\natexlab{b}}.

\bibitem[Niyogi et~al.(2008)Niyogi, Smale, and Weinberger]{niyogi2008finding}
Partha Niyogi, Stephen Smale, and Shmuel Weinberger.
\newblock Finding the homology of submanifolds with high confidence from random
  samples.
\newblock \emph{Discrete \& Computational Geometry}, 39:\penalty0 419--441,
  2008.

\bibitem[Odeyomi(2020)]{odeyomi2020learning}
Olusola~T. Odeyomi.
\newblock Learning the truth in social networks using multi-armed bandit.
\newblock \emph{IEEE Access}, 8:\penalty0 137692--137701, 2020.

\bibitem[Park et~al.(2014)Park, Park, goo Lee, and Jung]{park2014greedy}
Youngki Park, Sungchan Park, Sang goo Lee, and Woosung Jung.
\newblock Greedy filtering: A scalable algorithm for k-nearest neighbor graph
  construction.
\newblock In \emph{Database Systems for Advanced Applications: 19th
  International Conference, DASFAA 2014, Bali, Indonesia, April 21-24, 2014.
  Proceedings, Part I 19}, pages 327--341. Springer, 2014.

\bibitem[Perrault et~al.(2020)Perrault, Valko, and
  Perchet]{perrault2020covariance}
Pierre Perrault, Michal Valko, and Vianney Perchet.
\newblock Covariance-adapting algorithm for semi-bandits with application to
  sparse outcomes.
\newblock In \emph{Conference on Learning Theory}, pages 3152--3184. PMLR,
  2020.

\bibitem[Qin(2023)]{qin2023self}
Xiaoyu Qin.
\newblock \emph{Self-Adaptive Parameter Control Mechanisms in Evolutionary
  Computation}.
\newblock PhD thesis, University of Birmingham, 2023.

\bibitem[Reeve et~al.(2018)Reeve, Mellor, and Brown]{reeve2018k}
Henry Reeve, Joe Mellor, and Gavin Brown.
\newblock The k-nearest neighbour ucb algorithm for multi-armed bandits with
  covariates.
\newblock In \emph{Algorithmic Learning Theory}, pages 725--752. PMLR, 2018.

\bibitem[Rigollet and Zeevi(2010)]{rigollet2010nonparametric}
Philippe Rigollet and Assaf Zeevi.
\newblock Nonparametric bandits with covariates.
\newblock \emph{arXiv preprint arXiv:1003.1630}, 2010.

\bibitem[Riquelme et~al.(2018)Riquelme, Tucker, and Snoek]{riquelme2018neural}
Carlos Riquelme, George Tucker, and Jasper Snoek.
\newblock Deep bayesian bandits showdown: An empirical comparison of bayesian
  deep networks for thompson sampling.
\newblock \emph{arXiv preprint arXiv:1802.09127}, 2018.

\bibitem[Rudin(1964)]{rudin1964principles}
Walter Rudin.
\newblock \emph{Principles of Mathematical Analysis}, volume~3.
\newblock McGraw-hill New York, 1964.

\bibitem[Russac et~al.(2019)Russac, Vernade, and Capp{\'e}]{russac2019weighted}
Yoan Russac, Claire Vernade, and Olivier Capp{\'e}.
\newblock Weighted linear bandits for non-stationary environments.
\newblock \emph{Advances in Neural Information Processing Systems}, 32, 2019.

\bibitem[Russo and Roy(2013)]{russo2013eluder}
Daniel Russo and Benjamin~Van Roy.
\newblock Eluder dimension and the sample complexity of optimistic exploration.
\newblock \emph{Advances in Neural Information Processing Systems}, 26, 2013.

\bibitem[Sani et~al.(2012)Sani, Lazaric, and Munos]{sani2012risk}
Amir Sani, Alessandro Lazaric, and R{\'e}mi Munos.
\newblock Risk-aversion in multi-armed bandits.
\newblock \emph{Advances in Neural Information Processing Systems}, 25, 2012.

\bibitem[Schwartz et~al.(2017)Schwartz, Bradlow, and
  Fader]{schwartz2017customer}
Eric~M. Schwartz, Eric~T. Bradlow, and Peter~S. Fader.
\newblock Customer acquisition via display advertising using multi-armed bandit
  experiments.
\newblock \emph{Marketing Science}, 36\penalty0 (4):\penalty0 500--522, 2017.

\bibitem[Schwenk and Bengio(2000)]{schwenk2000boosting}
Holger Schwenk and Yoshua Bengio.
\newblock Boosting neural networks.
\newblock \emph{Neural computation}, 12\penalty0 (8):\penalty0 1869--1887,
  2000.

\bibitem[Shadkam(2022)]{shadkam2022parameter}
Elham Shadkam.
\newblock Parameter setting of meta-heuristic algorithms: A new hybrid method
  based on dea and rsm.
\newblock \emph{Environmental Science and Pollution Research}, 29\penalty0
  (15):\penalty0 22404--22426, 2022.

\bibitem[Shen et~al.(2015)Shen, Wang, Jiang, and Zha]{shen2015portfolio}
Weiwei Shen, Jun Wang, Yu-Gang Jiang, and Hongyuan Zha.
\newblock Portfolio choices with orthogonal bandit learning.
\newblock In \emph{Twenty-fourth International Joint Conference on Artificial
  Intelligence}, 2015.

\bibitem[Sherman and Morrison(1950)]{sherman1950adjustment}
Jack Sherman and Winifred~J. Morrison.
\newblock Adjustment of an inverse matrix corresponding to a change in one
  element of a given matrix.
\newblock \emph{The Annals of Mathematical Statistics}, 21\penalty0
  (1):\penalty0 124--127, 1950.

\bibitem[Shi et~al.(2023)Shi, Xiao, Pickard, Chen, and Chen]{shi2023deep}
Qicai Shi, Feng Xiao, Douglas Pickard, Inga Chen, and Liang Chen.
\newblock Deep neural network with linucb: A contextual bandit approach for
  personalized recommendation.
\newblock In \emph{Companion Proceedings of the ACM Web Conference 2023}, pages
  778--782, 2023.

\bibitem[Strang(2022)]{strang2022introduction}
Gilbert Strang.
\newblock \emph{Introduction to Linear Algebra}.
\newblock SIAM, 2022.

\bibitem[Turner et~al.(2021)Turner, Eriksson, McCourt, Kiili, Laaksonen, Xu,
  and Guyon]{turner2021bayesian}
Ryan Turner, David Eriksson, Michael McCourt, Juha Kiili, Eero Laaksonen, Zhen
  Xu, and Isabelle Guyon.
\newblock Bayesian optimization is superior to random search for machine
  learning hyperparameter tuning: Analysis of the black-box optimization
  challenge 2020.
\newblock In \emph{NeurIPS 2020 Competition and Demonstration Track}, pages
  3--26. PMLR, 2021.

\bibitem[Vaswani et~al.(2017)Vaswani, Shazeer, Parmar, Uszkoreit, Jones, Gomez,
  Kaiser, and Polosukhin]{vaswani2017attention}
Ashish Vaswani, Noam Shazeer, Niki Parmar, Jakob Uszkoreit, Llion Jones,
  Aidan~N. Gomez, {\L}ukasz Kaiser, and Illia Polosukhin.
\newblock Attention is all you need.
\newblock \emph{Advances in Neural Information Processing Systems}, 30, 2017.

\bibitem[Villar et~al.(2015)Villar, Bowden, and Wason]{villar2015multi}
Sof{\'\i}a~S. Villar, Jack Bowden, and James Wason.
\newblock Multi-armed bandit models for the optimal design of clinical trials:
  Benefits and challenges.
\newblock \emph{Statistical Science: A Review Journal of the Institute of
  Mathematical Statistics}, 30\penalty0 (2):\penalty0 199, 2015.

\bibitem[Yann et~al.(2010)Yann, Cortes, and Burges]{yann2010mnist}
LeCun Yann, Corinna Cortes, and CJ~Burges.
\newblock Mnist handwritten digit database.
\newblock \emph{ATT Labs}, 2010.

\bibitem[Zeng et~al.(2016)Zeng, Wang, Mokhtari, and Li]{zeng2016online}
Chunqiu Zeng, Qing Wang, Shekoofeh Mokhtari, and Tao Li.
\newblock Online context-aware recommendation with time varying multi-armed
  bandit.
\newblock In \emph{Proceedings of the 22nd ACM SIGKDD International Conference
  on Knowledge Discovery and Data Mining}, pages 2025--2034, 2016.

\bibitem[Zhang et~al.(2021)Zhang, Zhou, Li, and Gu]{zhang2021neural}
Weitong Zhang, Dongruo Zhou, Lihong Li, and Quanquan Gu.
\newblock Neural thompson sampling.
\newblock In \emph{International Conference on Learning Representations
  (ICLR)}, 2021.

\bibitem[Zhang(2019)]{zhang2019automatic}
Xiaojin Zhang.
\newblock Automatic ensemble learning for online influence maximization, 2019.
\newblock arXiv preprint arXiv:1911.10728.

\bibitem[Zhou et~al.(2020)Zhou, Li, and Gu]{zhou2020neural}
Dongruo Zhou, Lihong Li, and Quanquan Gu.
\newblock Neural contextual bandits with ucb-based exploration.
\newblock In \emph{International Conference on Machine Learning}, pages
  11492--11502. PMLR, 2020.

\bibitem[Zhou et~al.(2017)Zhou, Zhang, Xu, and Liang]{zhou2017large}
Qian Zhou, XiaoFang Zhang, Jin Xu, and Bin Liang.
\newblock Large-scale bandit approaches for recommender systems.
\newblock In \emph{Neural Information Processing: 24th International
  Conference, ICONIP 2017, Guangzhou, China, November 14-18, 2017, Proceedings,
  Part I 24}, pages 811--821. Springer, 2017.

\bibitem[Zhu et~al.(2022)Zhu, Wang, and Si]{zhu2022flexible}
Jinbiao Zhu, Dongshu Wang, and Jikai Si.
\newblock Flexible behavioral decision making of mobile robot in dynamic
  environment.
\newblock \emph{IEEE Transactions on Cognitive and Developmental Systems},
  15\penalty0 (1):\penalty0 134--149, 2022.

\end{thebibliography}
